\documentclass[letterpaper, 10 pt, conference]{ieeeconf}  %

\IEEEoverridecommandlockouts                              %

\usepackage{amsmath}
\usepackage{booktabs}
\usepackage{graphicx}
\usepackage[bookmarks=true,colorlinks]{hyperref}
\usepackage{multirow}

\newcommand{\ie}{\textit{i.e.}, }
\newcommand{\eg}{\textit{e.g.}, }
\newcommand{\mysubsection}[1]{\vspace{2mm}\noindent{\textbf{#1}}}
\DeclareMathOperator*{\argmax}{arg\,max}
\newcommand{\x}{\hphantom{1}}

\title{\LARGE \bf
Learning Pneumatic Non-Prehensile Manipulation with a Mobile Blower
}

\author{Jimmy Wu$^{1,2}$, Xingyuan Sun$^{1}$, Andy Zeng$^{2}$, Shuran Song$^{3}$, Szymon Rusinkiewicz$^{1}$, Thomas Funkhouser$^{1,2}$%
\thanks{$^{1}$Princeton University, $^{2}$Google, $^{3}$Columbia University}%
}

\begin{document}

\maketitle
\thispagestyle{empty}
\pagestyle{empty}

\begin{abstract}

We investigate pneumatic non-prehensile manipulation (i.e., blowing) as a means of efficiently moving scattered objects into a target receptacle. Due to the chaotic nature of aerodynamic forces, a blowing controller must (i) continually adapt to unexpected changes from its actions, (ii) maintain fine-grained control, since the slightest misstep can result in large unintended consequences (e.g., scatter objects already in a pile), and (iii) infer long-range plans (e.g., move the robot to strategic blowing locations). We tackle these challenges in the context of deep reinforcement learning, introducing a multi-frequency version of the spatial action maps framework. This allows for efficient learning of vision-based policies that effectively combine high-level planning and low-level closed-loop control for dynamic mobile manipulation. Experiments show that our system learns efficient behaviors for the task, demonstrating in particular that blowing achieves better downstream performance than pushing, and that our policies improve performance over baselines. Moreover, we show that our system naturally encourages emergent specialization between the different subpolicies spanning low-level fine-grained control and high-level planning. On a real mobile robot equipped with a miniature air blower, we show that our simulation-trained policies transfer well to a real environment and can generalize to novel objects.

\end{abstract}

\section{Introduction}

Pneumatic manipulation uses air pressure to exert forces on an
environment in order to achieve a goal state.
It is a form of dynamic non-prehensile manipulation where the action primitive is blowing, rather than pushing, rolling, etc.

In comparison to prehensile manipulation, pneumatic manipulation has
several advantages, particularly when working with lightweight, scattered objects.
First, it can move objects that are distant from the robot, effectively expanding the working volume of potential manipulations.
Second, it can move objects using the emitted air at speeds much greater than the maximum velocity of the robot itself.
Finally, it can manipulate objects with arbitrary shapes and materials without complicated grasp planning.
These features are some of the main reasons that landscapers use blowers rather than rakes for leaf cleanup.

In spite of these potential advantages, the use of pneumatic manipulation in robotics has been limited.
Prior works have designed systems that use air blowers to move objects along flat surfaces in industrial settings~\cite{laurent2015survey} or to 
control the trajectories of specific types of objects 
in highly controlled laboratory or industrial settings~\cite{becker2009automated,davis2008end,erzincanli1998design,ozcelik2002non}.
However, there is little to no work on utilizing blowers for dynamic
manipulation of arbitrary objects in more general settings.

\begin{figure}
\includegraphics[width=\columnwidth]{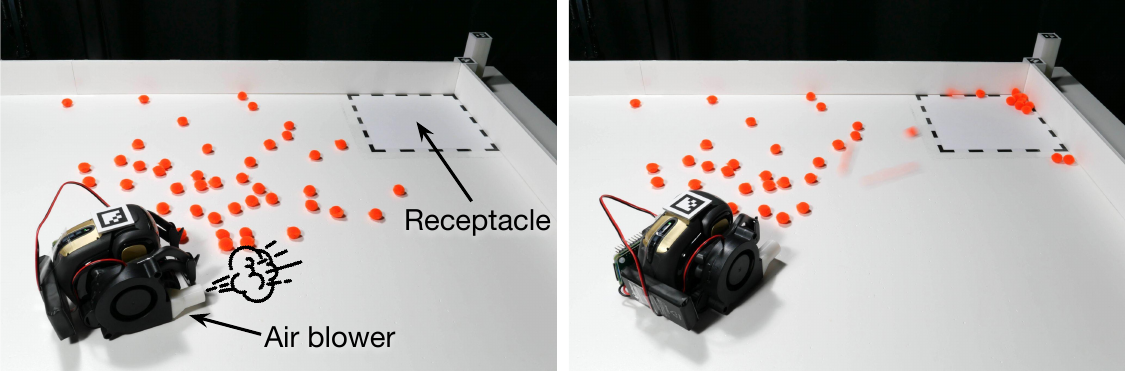}
\vspace{-5mm}
\caption{We train a mobile robot using deep reinforcement learning to clean up an environment using an air blower. The blower is used to efficiently move scattered objects (orange) into the target receptacle in the corner.}
\label{fig:teaser}
\vspace{-3mm}
\end{figure}

General pneumatic manipulation requires addressing several challenges.
First, due to the chaotic nature of aerodynamics (especially with obstacles), it is difficult to predict the effects of blowing on the state of the environment.
As a result, a blowing policy must continually sense the environment and adapt to unexpected changes in a closed-loop manner.
Similarly, fine-grained control is required to blow objects towards a target, and the slightest misstep can have large unintended consequences (\eg scattering objects already in a pile).
Finally, the complexity of achieving a target state using only air flow requires long-range planning.
For example, a policy might first move the robot to a strategic location with the blower off (so as not to scatter objects unintentionally), and only then turn the blower on to move objects towards the target.

\begin{figure*}
\includegraphics[width=\textwidth]{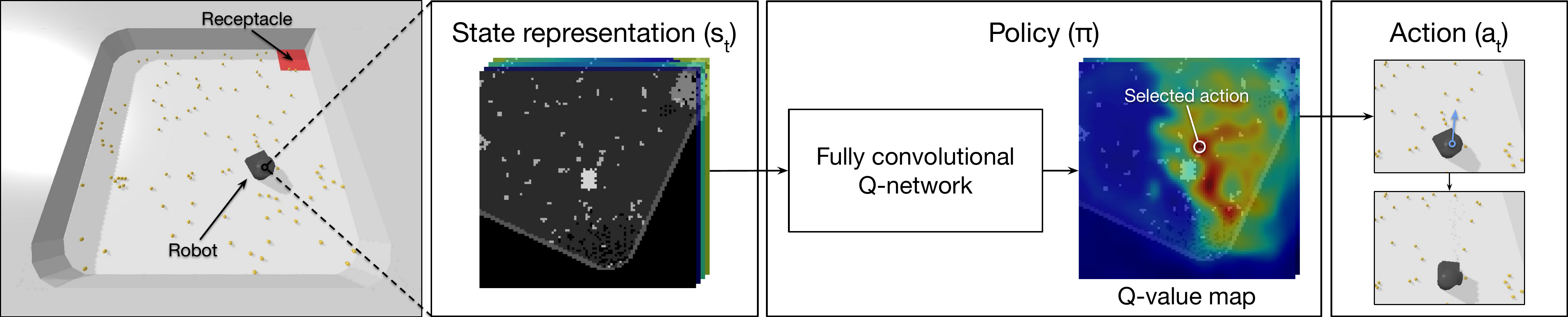}
\vspace{-5mm}
\caption{\textbf{Spatial action maps.} We use the spatial action maps framework~\cite{wu2020spatial} for our learned policies. Here we show a single step of policy execution. The policy network, which is implemented as a fully convolutional Q-network, takes in an image-based state representation and outputs a dense Q-value map, from which the best action is selected.}
\label{fig:spatial-action-maps}
\vspace{-3mm}
\end{figure*}

In this paper, we investigate whether a mobile robot equipped with a blower can learn to move scattered objects into a receptacle (Fig.~\ref{fig:teaser}).
We use deep reinforcement learning (deep RL) to train a robot to perform a sequence of movement and blowing actions.
In each step, the robot receives an image observation of its local environment and uses spatial action maps \cite{wu2020spatial} (Fig.~\ref{fig:spatial-action-maps}) to select its next action, where each action consists of a position to move to along with an action type (\eg \texttt{move-without-blowing} or \texttt{turn-while-blowing}).
We train all policies in simulation and test them directly in the real world.

In initial experiments, we found that the standard spatial action maps framework fails to learn effective policies in settings that require both long-term planning and fine-grained control.
To address this issue, we present a multi-frequency version of spatial action maps in which $n$ separate subpolicies are trained to operate together at different interleaved frequencies.
For example, if $n=2$, the top-level subpolicy executes an action, then the second-level subpolicy executes $k$ actions in a row, and the pattern repeats.
For each action, rewards are assigned to the subpolicy that chose it, and also to all subpolicies operating at lower frequencies.

In experiments with multi-frequency policies in both simulation and real-world settings (Fig.~\ref{fig:teaser}), we find that robots can indeed be trained to blow scattered objects into a receptacle, and that these robots are more efficient at the task than ones that push objects or use a single-frequency policy.
In real-world experiments, we find that our simulation-trained policies transfer well to the real environment, and can generalize to novel objects of different sizes and shapes.
For qualitative video results and code, please see our supplementary material at \url{https://learning-dynamic-manipulation.cs.princeton.edu}.

\section{Related Work}

\vspace{-2mm}\mysubsection{Pneumatic non-prehensile manipulation.}
Blowing is a form of pneumatic non-prehensile manipulation~\cite{lynch2003control,mason1999progress} as well as a form of dynamic manipulation~\cite{mason1993dynamic,wang2020swingbot,zeng2020tossingbot,ha2021flingbot,xu2022dextairity} in which the manipulator uses aerodynamic forces to effect change on the physical world.
Early work in this area used distributed air flow to manipulate planar objects for conveyance systems~\cite{konishi1999development, reed2004high} (\eg semiconductor wafers, biscuits) or to keep rectangular objects in equilibrium~\cite{luntz2001distributed}.
Air flow or pneumatic systems have also been used for sphere sorting~\cite{becker2009automated}, levitation~\cite{escano2005position,nordine1982aerodynamic}, manipulating sliced fruit and vegetables~\cite{davis2008end}, and manipulating non-rigid materials~\cite{erzincanli1998design} such as garments~\cite{ozcelik2002non,xu2022dextairity}, woven fabric~\cite{ozcelik2005examination}, and paper~\cite{biegelsen2000airjet}.
However, these systems have only been demonstrated in highly controlled laboratory or industrial settings, and do not support mobile manipulation of arbitrary objects scattered throughout an environment.

\mysubsection{Multi-frequency control.}
There has also been extensive research in incorporating compositional temporal structure for multi-frequency robot control: from constructing a hierarchical abstraction of control primitives, to combining them with reinforcement learning. Early work in this area includes Dynamic Movement Primitives (DMPs)~\cite{ijspeert2013dynamical,mulling2013learning,saveriano2021dynamic}, which use attractor dynamics to produce stable units of control that are sequenced or blended together to perform downstream tasks with imitation learning. DMPs have also since been extended and used within hierarchical reinforcement learning (RL) using the options framework~\cite{daniel2012hierarchical, stulp2012reinforcement,sutton1999between}, where they are formulated as pretrained low-level skills that are composed hierarchically by having a high-level policy choose between primitive actions or pretrained skills~\cite{tessler2017deep}.

\mysubsection{Hierarchical reinforcement learning.}
Within hierarchical RL, there is also a class of compositional methods in which a high-level policy issues commands to be executed by a low-level policy.
The commands can be specified by learned latent representations \cite{heess2016learning,haarnoja2018latent,wang2021hierarchical}), low-level control parameters \cite{schaal2006dynamic,stulp2011hierarchical,bahl2020neural,bahl2021hierarchical}, or subgoals, which can be hand-designed~\cite{kulkarni2016hierarchical}, task-specific locations~\cite{peng2017deeploco,faust2018prm,wahid2019long,wang2020model,nachum2020multi,li2020hrl4in,xiali2020relmogen}, target object states~\cite{stolle2002learning,levy2017learning,nachum2018data}), or target states in a learned latent space~\cite{morimoto2001acquisition,vezhnevets2017feudal}.
In contrast to these works, our multi-frequency system uses no explicit skills, commands, or subgoals.
Our subpolicies operate at different frequencies, but they select from the same set of primitive actions, and there is no explicit communication from the high-level to low-level.
Nevertheless, division of labor between subpolicies in our system emerges serendipitously during training.

Different approaches also differ in their training regimens. Systems can be trained stage-wise, where the controllers operating at the highest frequencies (\eg for control primitives) are first trained and then frozen when training the high-level policies~\cite{florensa2017stochastic,tessler2017deep,peng2017deeploco,haarnoja2018latent,nachum2020multi,wang2020model, camacho2021reward}, or all controllers and policies can be trained simultaneously~\cite{frans2017meta,levy2017learning,li2020hrl4in,sun2022fully}.
While the high-level policy typically gets its learning signal from the environment, the low level controller can be pretrained with custom proxy rewards such as robot velocity or grasp success~\cite{florensa2017stochastic,haarnoja2018latent}, intrinsic rewards for successfully following goals~\cite{kulkarni2016hierarchical,vezhnevets2017feudal,nachum2018data,nachum2020multi}, or even no explicit reward~\cite{eysenbach2018diversity}.
In contrast, we train all subpolicies together at the same time, and rely on the different frequencies that they execute with, as well as the different
rewards they receive, to learn specialization as an emerging behavior.

\section{Method}

We investigate how to train a robot equipped with a blower to move scattered objects into a target receptacle.
At each execution step (Fig. \ref{fig:spatial-action-maps}), the robot receives a new image observation of its local environment, uses online mapping to generate a state representation, and then selects the next action using spatial action maps \cite{wu2020spatial}.
Unlike in previous systems, the learned policy is multi-frequency:
it passes control of execution through an interleaved set of subpolicies executing at different frequencies.
These subpolicies are all trained together end-to-end, as described in Sec.~\ref{sec:multi-frequency-rl}.

\subsection{Reinforcement Learning Framework}

We model our task using a Markov decision process (MDP).
Given state $s_t$ at time $t$, the agent follows policy $\pi$ and takes action $a_t=\pi(s_t)$, receiving reward $r_t$ while entering new state $s_{t+1}$.
We train our policies using deep Q-learning (DQN)~\cite{mnih2015human}, where the goal is to find an optimal policy $\pi^*$ that selects actions to maximize the discounted sum of future rewards
$Q(s_t,a_t)=\sum_{i=t}^{\infty}\gamma^{i-t} r_i$,
typically referred to as the Q-function.
We follow DQN and approximate the Q-function using a neural network (Q-network), and use a policy that greedily selects actions to maximize the Q-function:
$\pi(s_t) = \argmax_{a_t} Q_\theta(s_t,a_t)$,
where $\theta$ refers to the parameters of the Q-network.
We use the double DQN variant~\cite{van2016deep} with smooth L1 loss.
Formally, at each training iteration $i$, we minimize
\resizebox{\columnwidth}{!}{$\mathcal{L}_i = |r_t + \gamma Q_{\theta_i^{\text{--}}}(s_{t+1},\argmax_{a_{t+1}}{Q_{\theta_i}(s_{t+1},a_{t+1})})-Q_{\theta_i}(s_t,a_t)|$},
where $(s_t,a_t,r_t,s_{t+1})$ is a transition uniformly sampled at random from the replay buffer, and $\theta^-$ refers to the parameters of the DQN target network.

\begin{figure}
\includegraphics[width=\columnwidth]{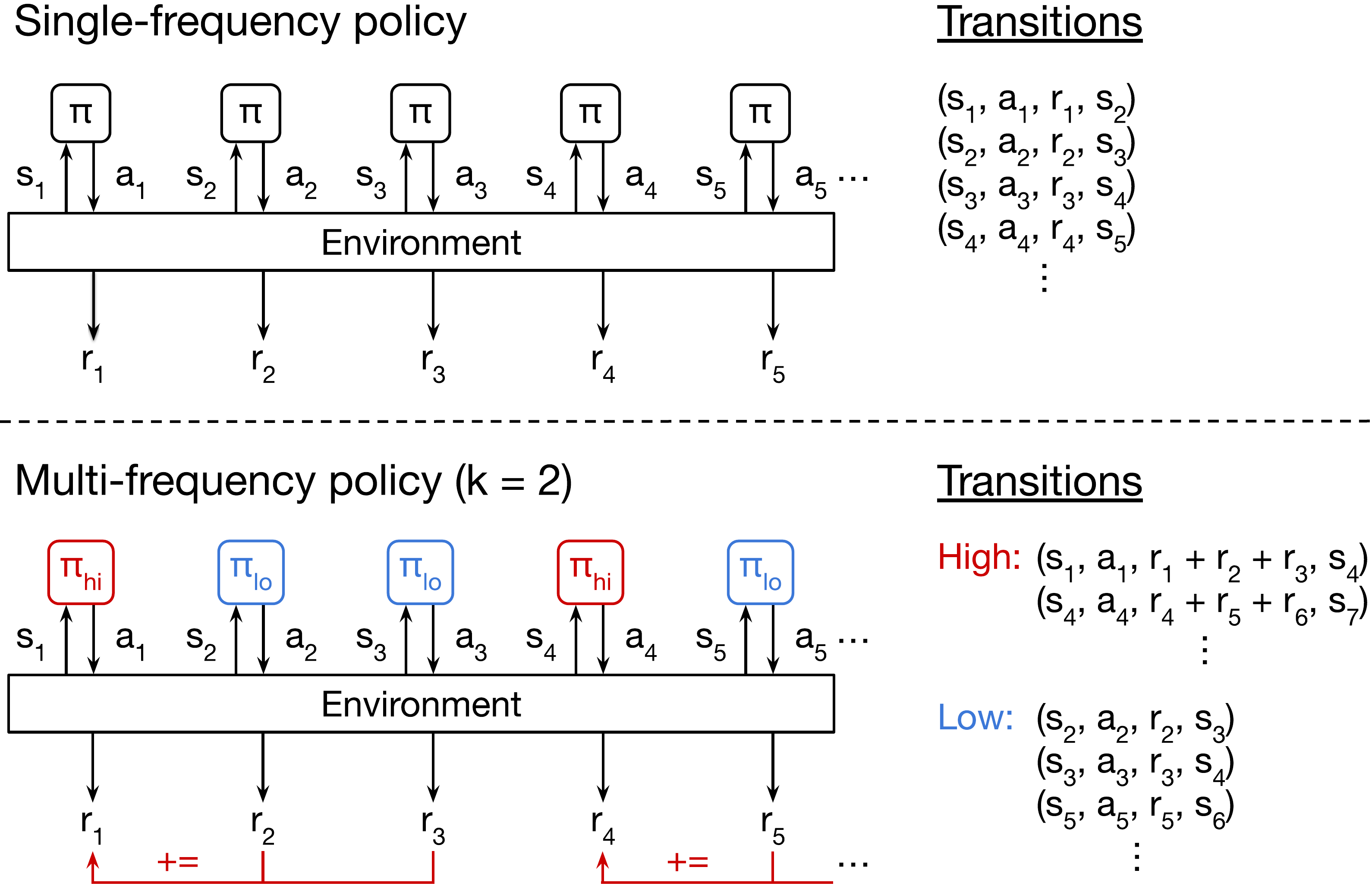}
\vspace{-5mm}
\caption{\textbf{Single-frequency vs.\ multi-frequency.}
In our multi-frequency framework, the low-level (high-frequency) subpolicy runs for $k$ consecutive steps after every high-level (low-frequency) subpolicy step. There is no explicit communication between subpolicies. However, rewards from the low-level steps are accumulated and given to the high level.}
\label{fig:multi-frequency}
\vspace{-3mm}
\end{figure}

\begin{figure*}
\begin{center}
\setlength\tabcolsep{1pt}
\begin{tabular}{ccccc}
\includegraphics[width=0.195\textwidth]{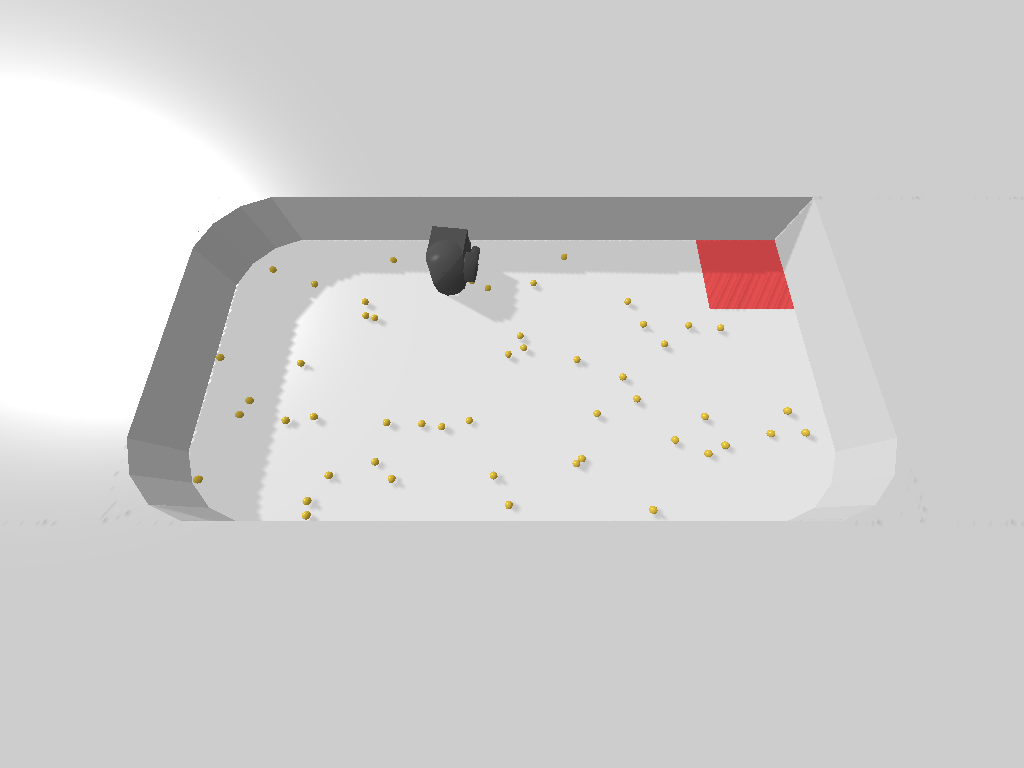} &
\includegraphics[width=0.195\textwidth]{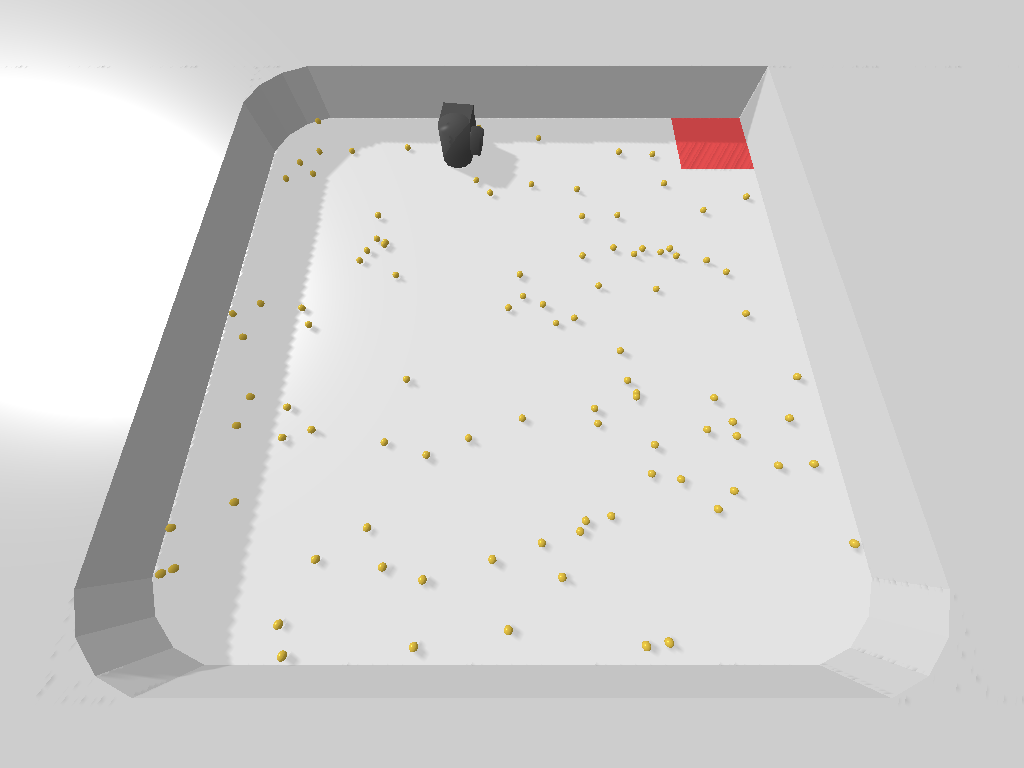} &
\includegraphics[width=0.195\textwidth]{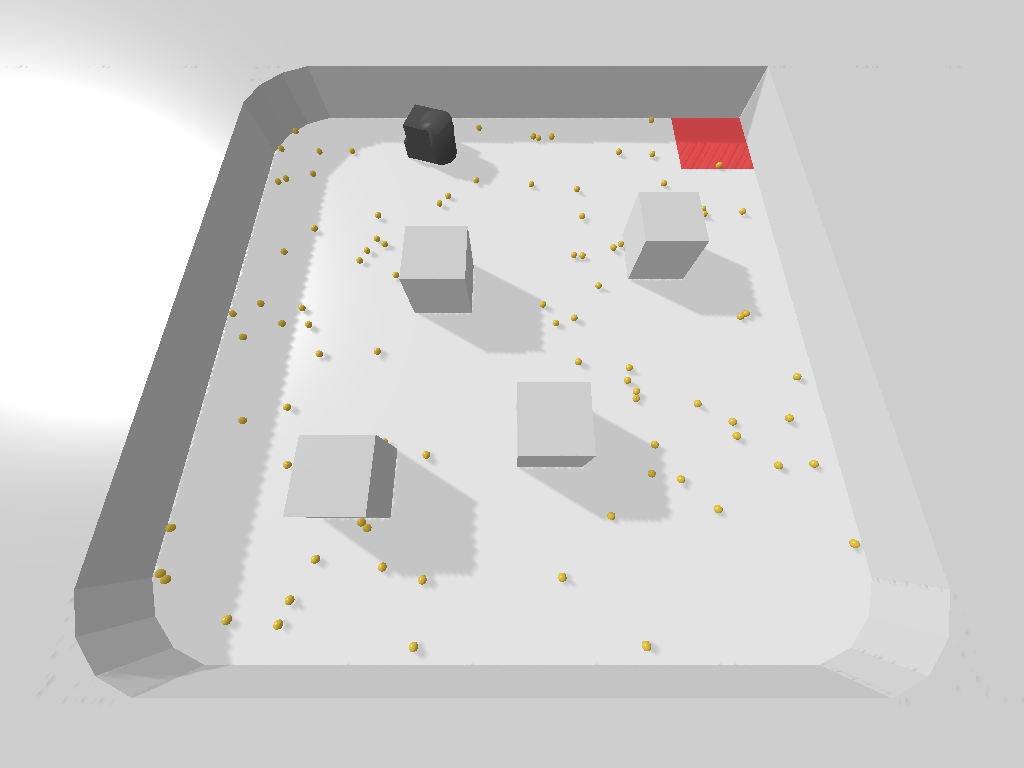} &
\includegraphics[width=0.195\textwidth]{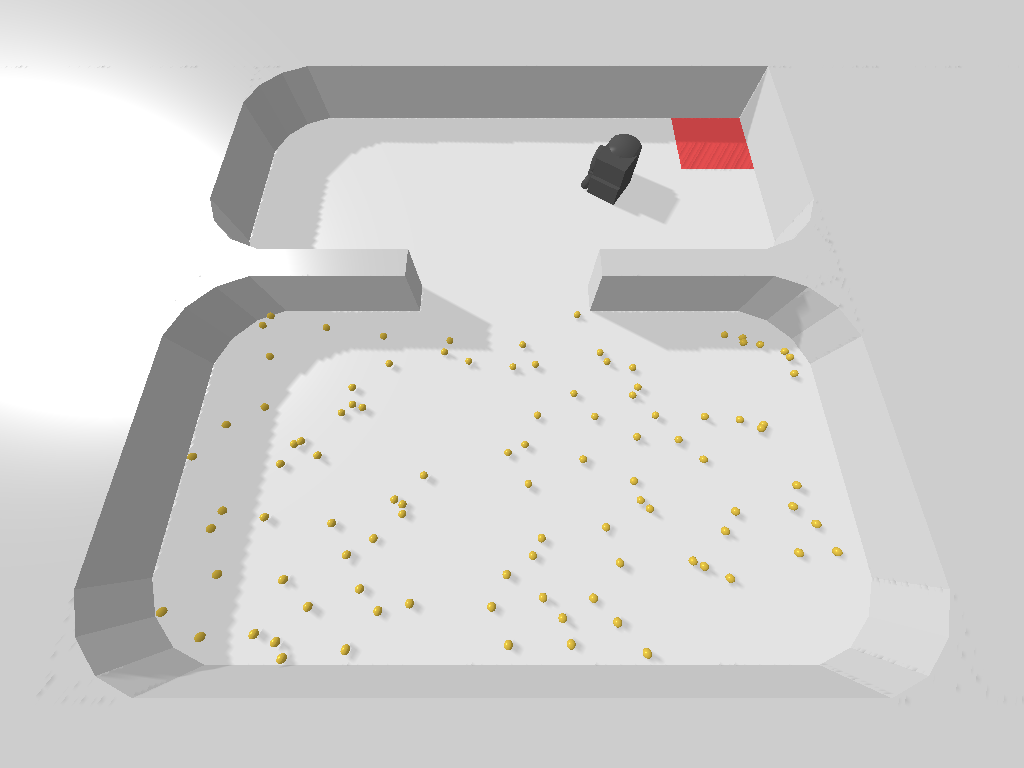} &
\includegraphics[width=0.195\textwidth]{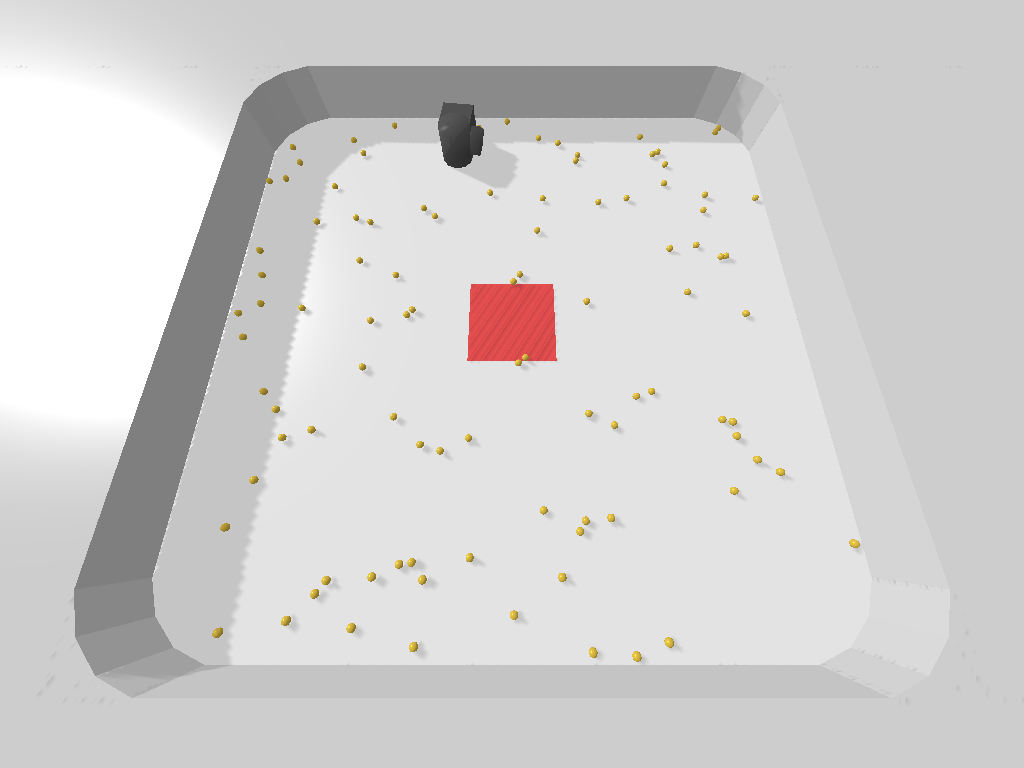} \\
\small{SmallEmpty} & \small{LargeEmpty} & \small{LargeColumns} & \small{LargeDoor} & \small{LargeCenter} \\
\end{tabular}
\end{center}
\vspace{-2mm}
\caption{\textbf{Environments.} We run experiments across several different environments. The first four involve moving objects to a receptacle in the corner, possibly around obstacles (LargeColumns and LargeDoor), while the fifth (LargeCenter) requires moving objects to a receptacle in the center.}
\label{fig:envs}
\vspace{-3mm}
\end{figure*}

\subsection{Multi-Frequency Reinforcement Learning}
\label{sec:multi-frequency-rl}

We formulate multi-frequency reinforcement learning as a multi-level optimization problem, where each level has its own execution frequency (Fig.~\ref{fig:multi-frequency}).
We impose a temporal frequency where $k$ low-level steps are taken per high-level step.
For example, a simple two-level policy might execute 4 low-level steps for every high-level step ($k = 4$).
For multi-level policies with more than two levels, we specify $k$ between every pair of consecutive frequency levels.
A sequence of actions begins with a high-level step, followed by $k$ steps at the next level, and so on, before returning control back to the high level.

We model each frequency level of the policy using a separate MDP, and train a separate subpolicy for each.
There is no explicit communication between levels, but we encourage the levels to work together via a learning signal that passes rewards from lower to higher levels.
In other words, each subpolicy gets all of the rewards given to higher-frequency (lower-level) policies since the last time it was run.

For example, suppose there are two levels \texttt{lo} and \texttt{hi}, whose Q-network parameters are $\theta_{\texttt{lo}}$ and $\theta_{\texttt{hi}}$. If we were executing the low-level policy, at time $t$ and in state $s_t$, we would take action $a_t=\argmax_{a} Q_{\theta_\texttt{lo}}(s_t, a)$, enter new state $s_{t+1}$, and receive reward $r_t$.
If we were instead executing the high-level policy, we would first take one action following the high-level policy, \ie $a_t=\argmax_{a} Q_{\theta_\texttt{hi}}(s_t, a)$.
We would then take $k$ consecutive actions following the low-level policy, \ie $a_{t + i}=\argmax_{a} Q_{\theta_\texttt{lo}}(s_{t + i}, a)$ for $i = 1$ through $k$.
We would end up in state $s_{t + k + 1}$, and the high-level policy would receive the reward of all $k + 1$ steps, $\sum_{i = 0}^{k} r_{t + i}$.
This includes not only the reward for its own action, but also the low-level rewards that the action enabled.

\subsection{State and Action Representations}

Both our state and action representations are based on spatial action maps~\cite{wu2020spatial}, which have been shown to achieve strong performance for discrete vision-based mobile manipulation tasks.
The state representation consists of an egocentric local overhead map along with additional spatially-aligned maps that are useful to the agent.
These local maps, represented as image channels, are oriented with the agent in the center, and in our system consist of:
(i) an overhead map of the environment,
(ii) an agent map encoding the pose of the agent,
(iii) a map encoding the shortest path distance to the receptacle, and
(iv) a map encoding the shortest path distance from the agent.

In particular, these local maps are constructed by taking oriented crops of global maps, which the agent gradually builds as it moves around the environment.
The agent starts with a blank map and must learn to seek out unexplored regions of the environment.
As in Wu et al.~\cite{wu2020spatial}, we simulate this online mapping process by affixing a simulated forward-facing RGB-D camera to the agent in simulation.
These cameras capture partial observations of the environment, which are fused with prior observations to construct a global overhead map and occupancy map.
Note that the partial observability means the maps may contain outdated information, and the agent must learn to handle this uncertainty.

Our action space is a multi-channel pixel map, where each pixel represents a different action conditioned on its corresponding location in the environment. The blowing robot uses two channels, one representing pure movement to a navigational endpoint (\texttt{move-without-blowing}), and a second representing (depending on the experiment) either turning (without translation) towards the selected location or moving to the location, all while blowing (\texttt{turn-while-blowing} and \texttt{move-while-blowing}). The pushing robot uses only the first channel, and serves as one of our baselines. As in Wu et al.~\cite{wu2020spatial}, all pixel maps are spatially aligned with the state representation, which allows learned policies to anchor their Q-values on visual features of the scene.
We choose an action greedily by computing the $\argmax$ across all channels in the action space, and then execute it using one of the following movement primitives: (i) a turning primitive which simply turns in place to face the selected location, and (ii) moving primitive which follows the shortest path (computed using the agent's occupancy map) to the selected location.

\subsection{Implementation Details}

\mysubsection{Simulation environments.} We train all policies in environments built with PyBullet~\cite{coumans2021pybullet}, where we simulate blowing by shooting small invisible spherical ``wind'' particles out of the blower. The particles are able to move objects in the environments using PyBullet's contact dynamics. We tuned the simulation to qualitatively match the behavior of the air blower used on the real robot.

\mysubsection{Network architecture.} We use a ResNet-18~\cite{he2016deep} backbone for our Q-function, which we transformed into a fully convolutional network~\cite{long2015fully} by removing the AvgPool and fully connected layers and adding three $1 \times 1$ convolution layers interleaved with bilinear upsampling layers. We include BatchNorm layers after all convolution layers.

\mysubsection{Rewards.} We give three types of rewards for training reinforcement learning: (i) a success reward of $+1$ whenever an object enters the receptacle, (ii) distance-based partial rewards or penalties whenever an object is moved closer to or further away from the receptacle, and (iii) a collision penalty of $-0.25$ whenever the robot hits a wall or an obstacle.

\mysubsection{Exploration.} We run a random policy at the beginning of training for a small number of environment timesteps (1/40 of total) to fill up the replay buffer with some initial transitions before we begin training the policy. We follow DQN and use $\epsilon$-greedy exploration, with the exploration fraction linearly annealed from 1 to 0.01 during the first 1/10 of the training.

\mysubsection{Training details.} We train all policies using SGD for 20k iterations, with batch size 32, learning rate 0.01, momentum 0.9, and weight decay 0.
Gradient norms are clipped to 100 during training.
We use a replay buffer of size 10k, a constant discount factor of $\gamma=0.75$, and a target network update frequency of 1,000 SGD iterations.
For single-frequency policies, we use a train frequency of 4, in which the environment is run for 4 timesteps for every SGD iteration.
For multi-frequency policies, we set the train frequency to be the number of lowest level steps per highest level, which is $k$ for 2-level policies and $k^2$ for 3-level policies. This ensures that the training pace is one SGD step per high-level subpolicy step.
Episodes end after all objects have been removed from the environment, or after 100 consecutive steps in which no object has been moved into the receptacle.
Training takes approximately 6 hours on a single Nvidia Titan Xp GPU, but can vary depending on the train frequency.

\section{Experiments}

We ran a series of experiments to evaluate how well our proposed methods learn to complete the task in both simulated and real-world environments.
Unless otherwise specified, experiments were conducted with robots that have a choice between two types of possible actions at each step: \texttt{move-without-blowing} and \texttt{turn-while-blowing}.
These actions were available equally to all levels of the policy.

Experiments were run in five different environments (see Fig.~
\ref{fig:envs}), each with its own unique challenges, such as moving objects around randomly placed obstacles (LargeColumns and LargeDoor) or towards a center receptacle (LargeCenter).
We initialize the episodes with 50 objects for SmallEmpty and 100 objects for the others.
For LargeColumns, we initialize each episode by placing a random number of columns (between 1 and 6) at random locations in the environment.
For LargeDoor, the doorway position is randomly offset each episode in both the x and y directions.

We trained 5 policies (5 runs) for each experiment and then ran each policy for 20 evaluation episodes.
Each episode is evaluated by counting the number of objects moved into the receptacle within a fixed time cutoff.
The time cutoff is different for each environment but kept consistent within the environment.
With this metric, we can compare the time efficiency of different methods.
We average the performance across evaluation episodes and report the mean and standard deviation across the 5 training runs.

\subsection{Simulation Experiments}

\begin{figure*}
\begin{center}
\setlength\tabcolsep{1pt}
\begin{tabular}{ccccc}
\includegraphics[width=0.195\textwidth]{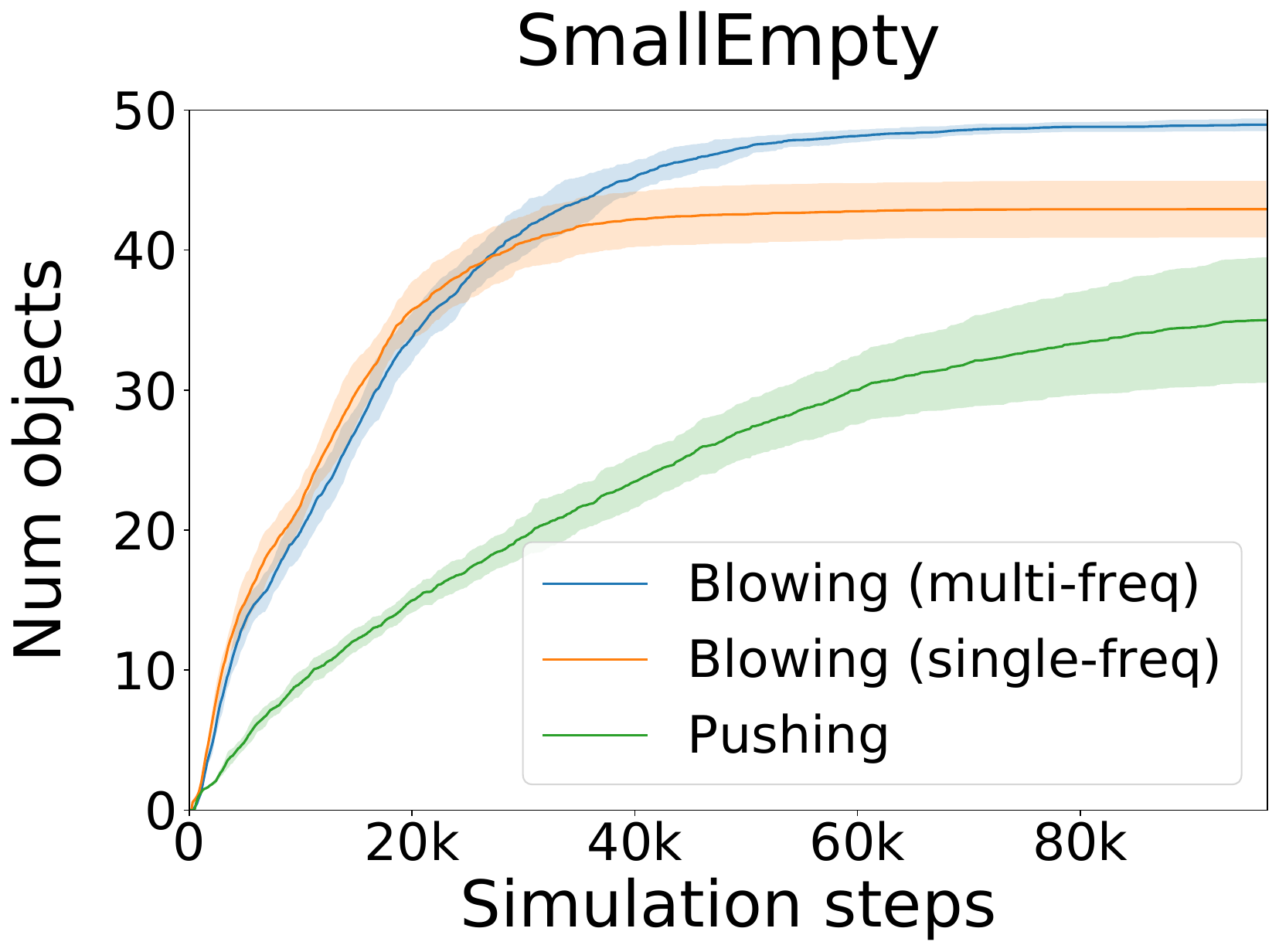} &
\includegraphics[width=0.195\textwidth]{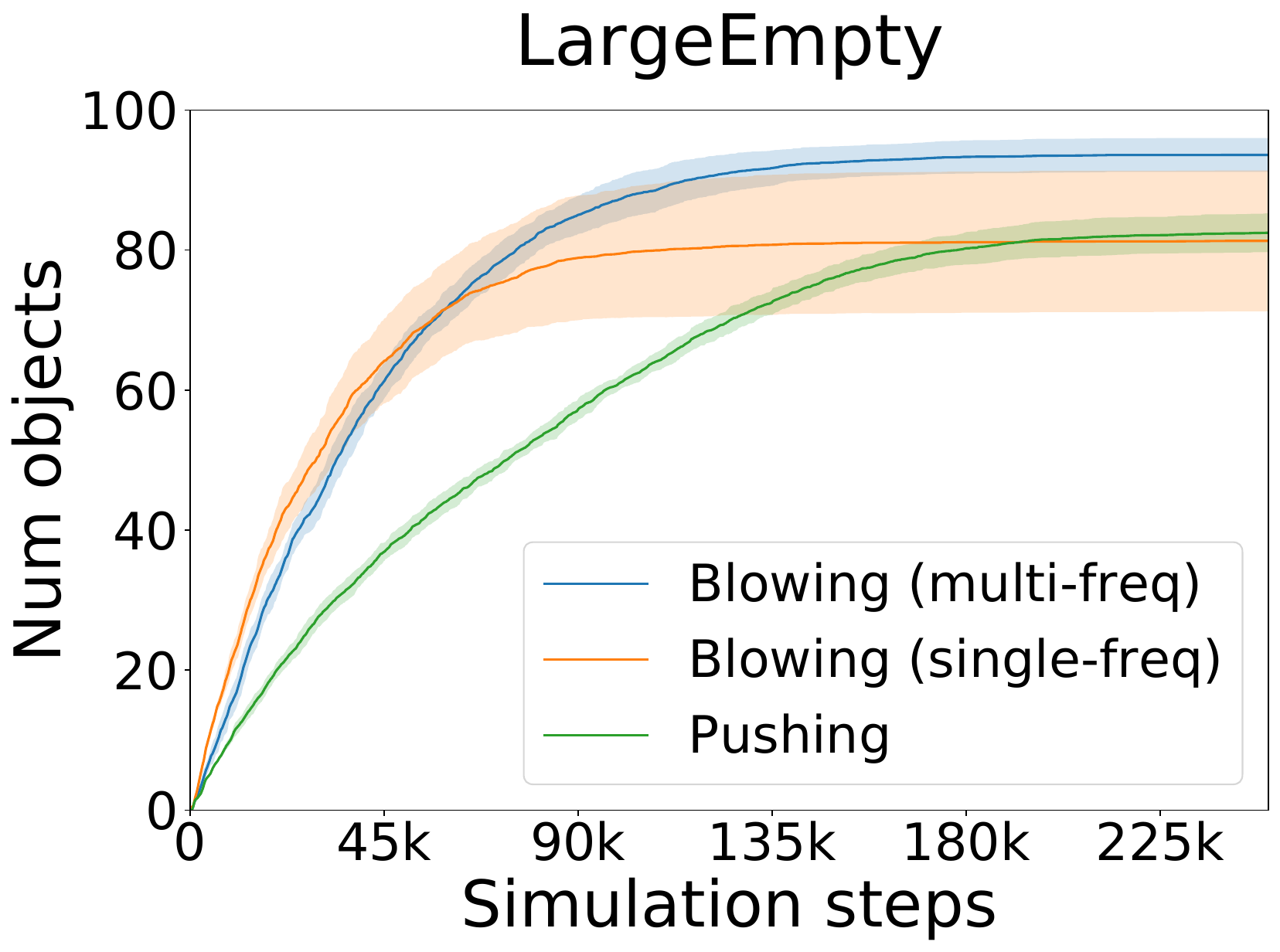} &
\includegraphics[width=0.195\textwidth]{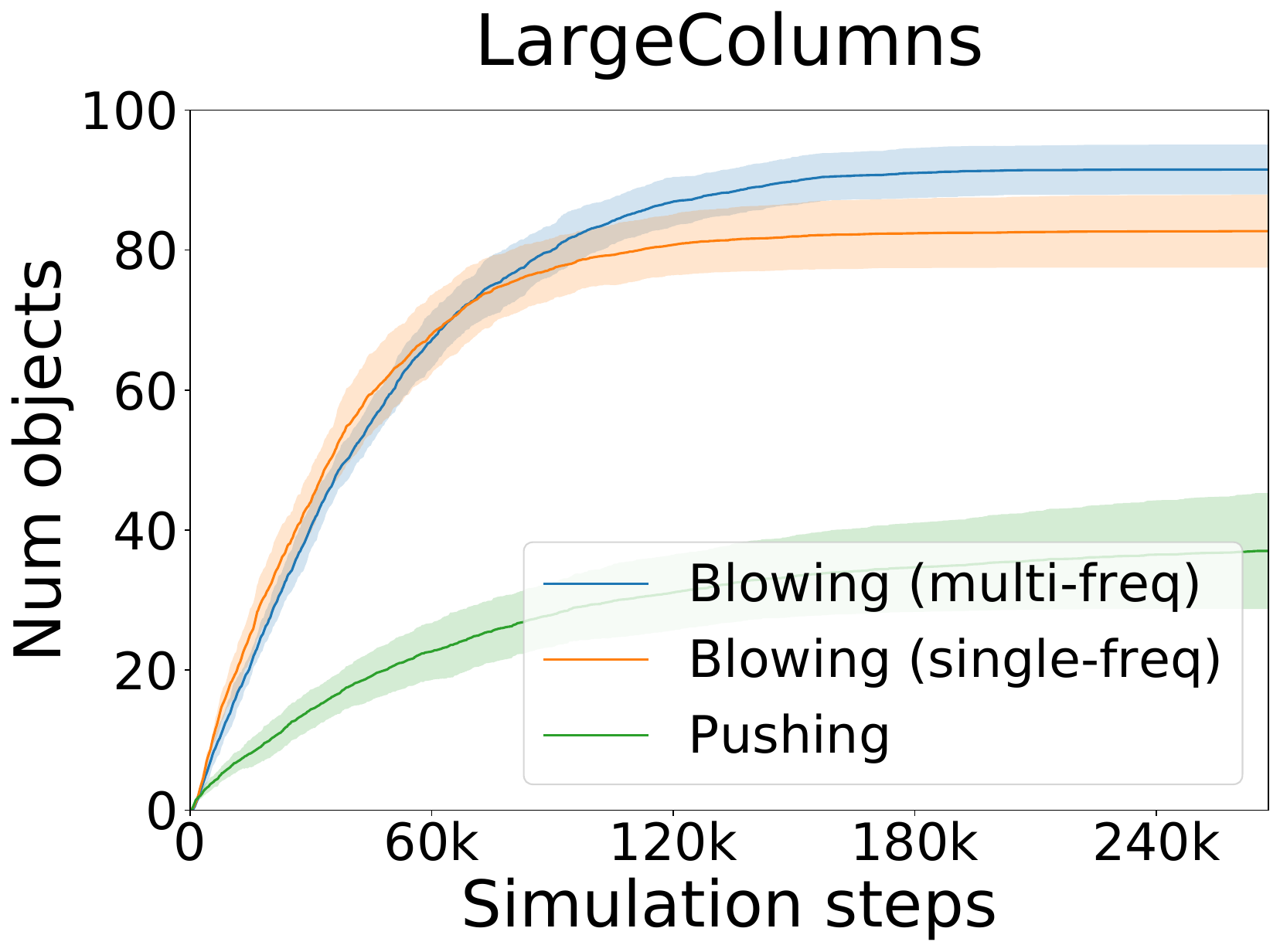} &
\includegraphics[width=0.195\textwidth]{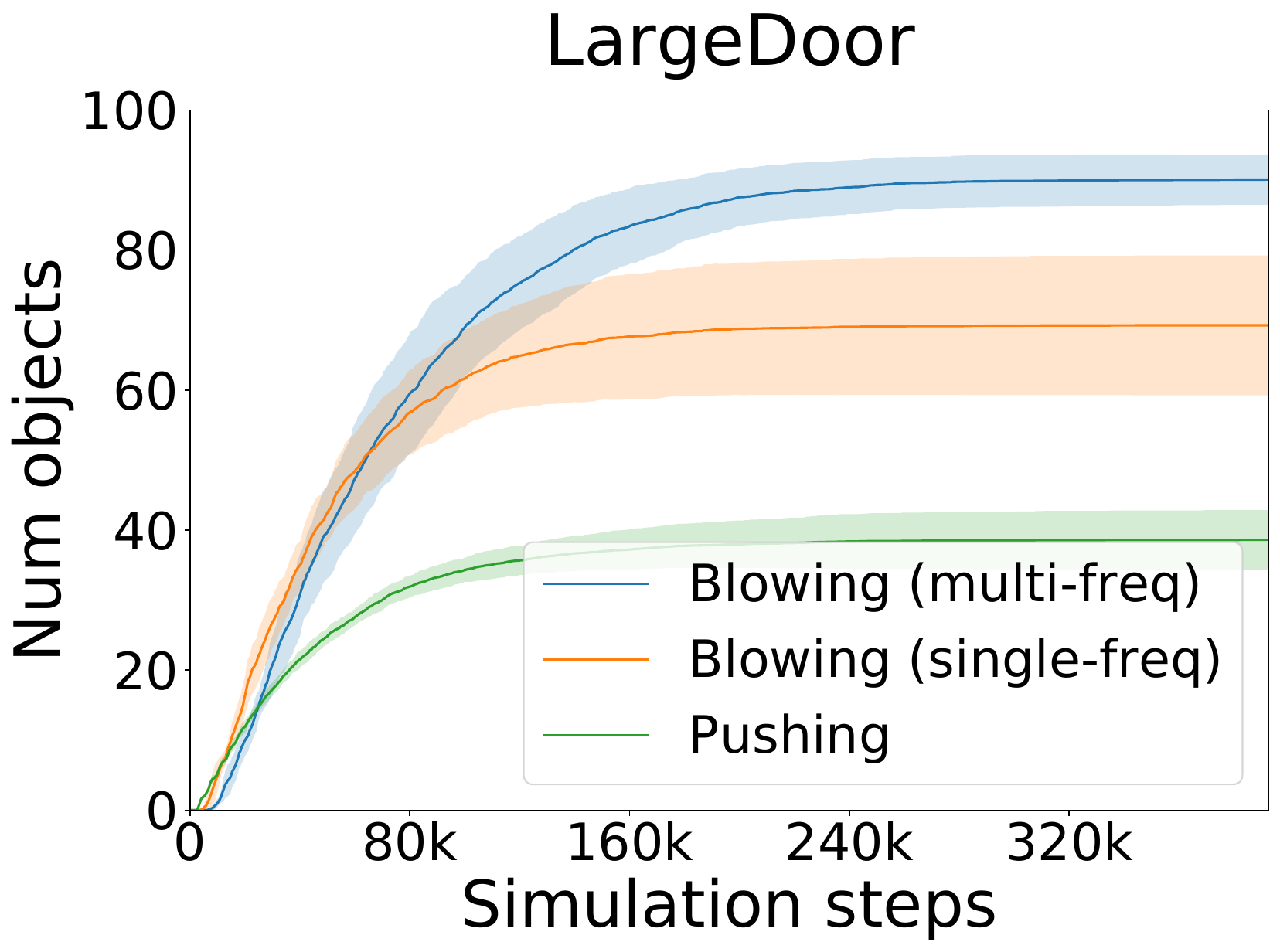} &
\includegraphics[width=0.195\textwidth]{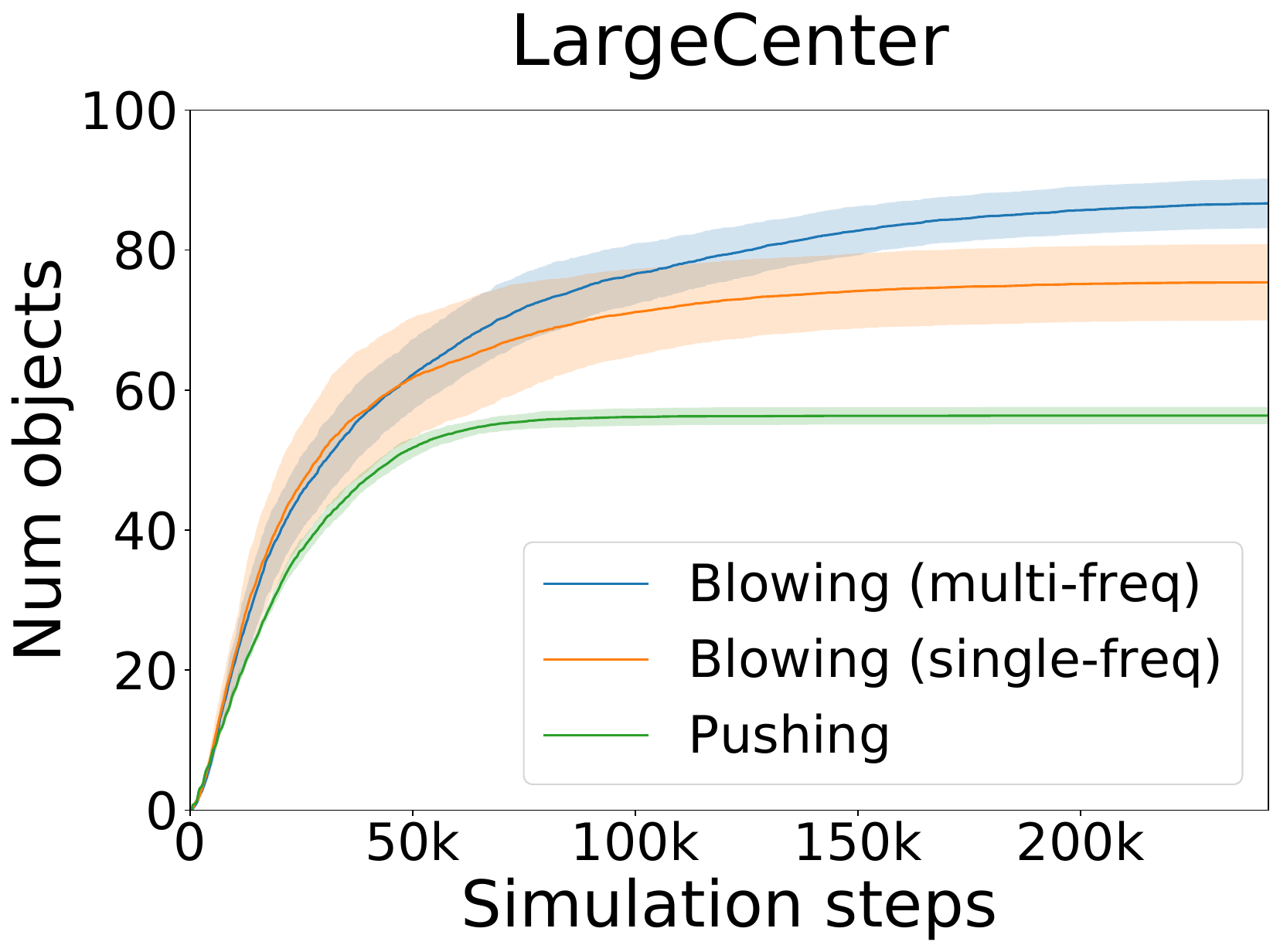} \\
\end{tabular}
\end{center}
\vspace{-3mm}
\caption{\textbf{Blowing vs.\ pushing efficiency.}
These evaluation curves show the number of objects moved into the receptacle over an episode in each of the five environments we studied. We can see from the steepness of the curves that blowing (blue and orange) is much more efficient than pushing (green).}
\label{fig:blowing-pushing-curves}
\vspace{-3mm}
\end{figure*}

\begin{figure}
\begin{center}
\setlength\tabcolsep{1pt}
\begin{tabular}{cc}
\includegraphics[width=0.49\columnwidth]{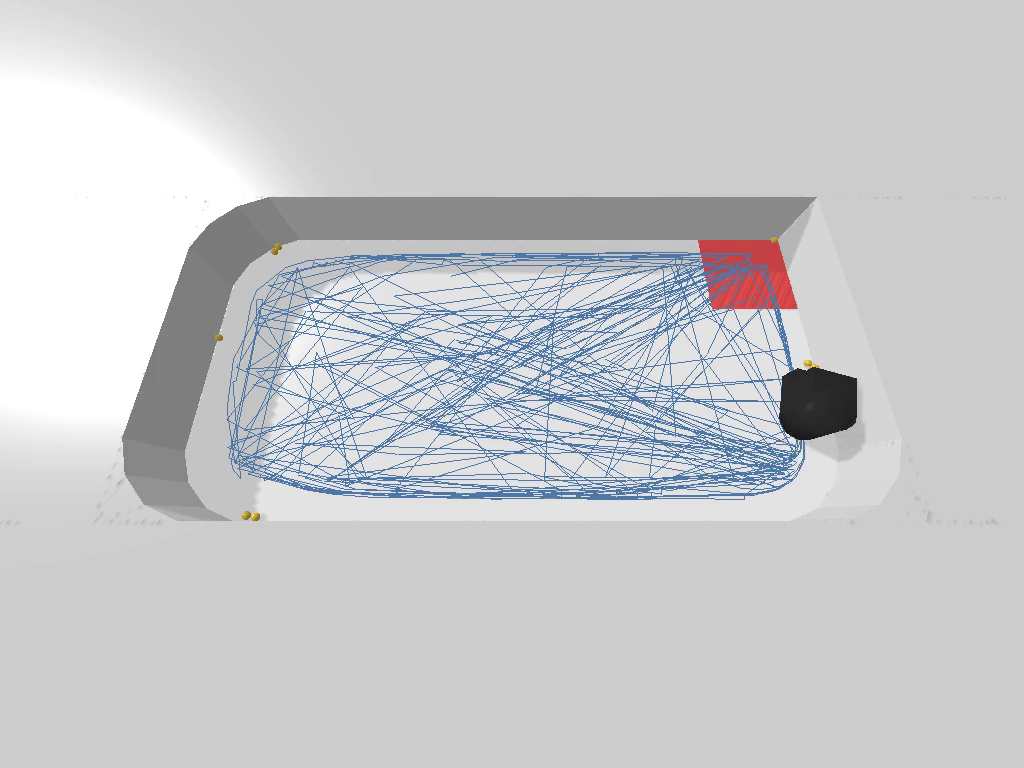} &
\includegraphics[width=0.49\columnwidth]{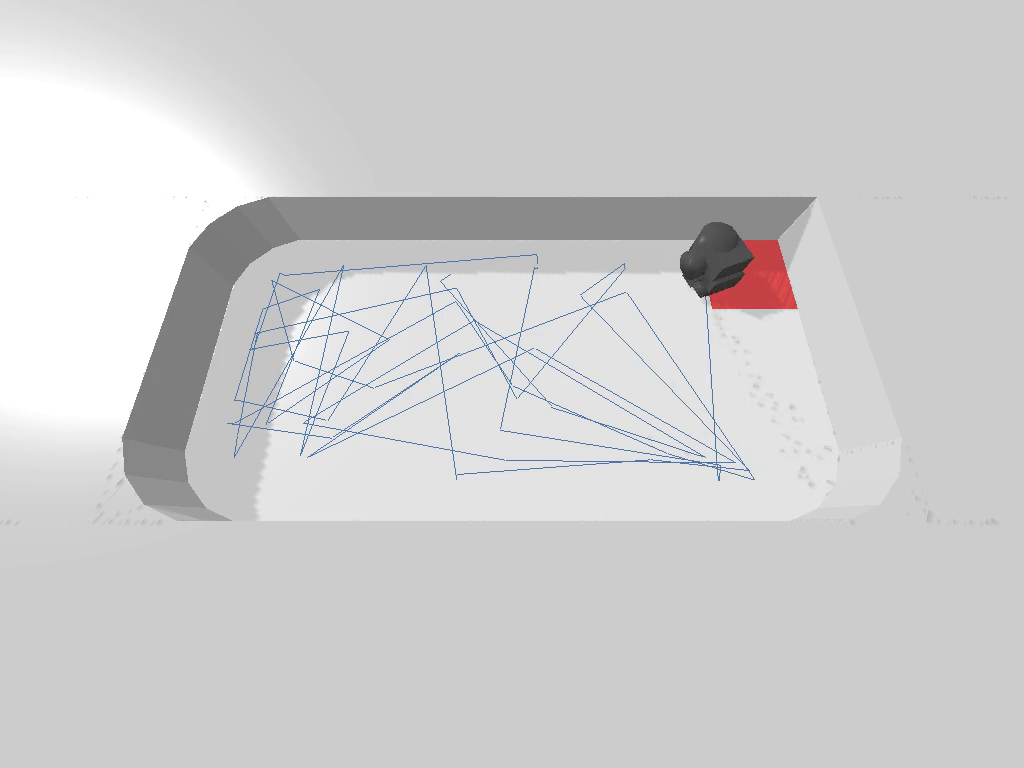} \\
\small{Pushing} & \small{Blowing} \\
\end{tabular}
\end{center}
\vspace{-2mm}
\caption{\textbf{Blowing vs.\ pushing trajectories.} The blue lines show movement trajectories over an evaluation episode.
We can clearly see that the blowing robot is much more efficient than the pushing robot.}
\label{fig:blowing-pushing-trajectories}
\vspace{-3mm}
\end{figure}

\mysubsection{Blowing vs.\ pushing.} Our first experiment tests the performance of robots that can blow objects in comparison to ones that can only push them.
We train separate policies for each with the same framework, but give the pushing robots $3\times$ more training steps than blowing to give them every advantage.
We also disable the collision penalty for the pushing robots so as to not disadvantage pushing along walls.
The results in Fig.~\ref{fig:blowing-pushing-curves} show plots of the number of objects moved into the receptacle at different numbers of simulation steps (elapsed time) within an episode.
They show that blowing is significantly more efficient than pushing. %
In SmallEmpty (left plot), for example, the blowing robots ``clean up'' objects around twice as fast as the pushing robots. The reason for the difference can be seen clearly in the 
movement trajectories visualized in Fig.~\ref{fig:blowing-pushing-trajectories}.
The blowing robot hardly has to move at all, while the pushing robot has to move into every corner and along every wall to move the objects.
We also see that the pushing robot tends to ignore objects near the corners as their small size makes them difficult to push, whereas the blowing robot has no trouble with them.
The advantages of blowing over pushing are even greater in the other environments.
In LargeColumns and LargeDoor, the blower manipulates objects around obstacles with ease, while the pusher needs to carefully push objects along each individual obstacle.
In LargeCenter, the blower can easily handle objects sitting along the walls and corners by blowing them out, whereas the pushing robot has a much harder time moving them back out towards the center.

\mysubsection{Multi-frequency vs.\ single-frequency.} Our second experiment compares the proposed two-level multi-frequency policy with a single-frequency one.
Both policies are trained with a frequency of 4 environment steps per SGD step to make the comparison fair.
We show the performance difference in Fig.~\ref{fig:blowing-pushing-curves} and Tab.~\ref{tab:multi-frequency}.
We find that both policies have similar initial performance, but the flat policies get stuck near the ends of episodes, repeatedly blowing at empty space.
We believe that this is due to inefficient learning caused by the imbalance in frequencies for effective movement and blowing actions: in good policies, blowing is more frequent than movement.
The multi-frequency policies use separate subpolicies with different execution frequencies, and are able to learn actions matched to those frequencies during training, thus resulting in better performance.
This emergent specialization within the two subpolicies of a trained multi-frequency policy can be seen in 
Tab.~\ref{tab:emergent-specialization}, which shows representative action frequencies for the blowing robot in the SmallEmpty environment.
We see that the high level learns to execute \texttt{move-without-blowing} actions predominantly (89.2\%), whereas the low level learns to execute mainly \texttt{turn-while-blowing} actions (82.3\%).
This specialization corresponds to a strategy of repeatedly moving to a good location and then turning/blowing from there for a while (see videos in supplementary materials).
While this strategy seems intuitive, it was not hand-crafted: both the low and high levels have the same sets of possible actions available to them.
The only difference is that the low level is executed more frequently, and when combined with our reward structure, this difference leads to the emerging behavior of specialization.

\begin{table}[h]
\vspace{-1mm}
\caption{Multi-frequency vs.\ single-frequency}
\vspace{-4mm}
\begin{center}
\begin{tabular}{l|cc}
\toprule
Environment & Single-frequency & Multi-frequency \\
\midrule
SmallEmpty & 42.87 $\pm$ \x2.01 & \textbf{48.52 $\pm$ 0.37} \\
LargeEmpty & 81.07 $\pm$ 10.07 & \textbf{93.09 $\pm$ 2.29} \\
LargeColumns & 82.40 $\pm$ \x4.95 & \textbf{91.01 $\pm$ 3.58} \\
LargeDoor & 69.12 $\pm$ \x9.82 & \textbf{89.58 $\pm$ 3.75} \\
LargeCenter & 75.35 $\pm$ \x5.45 & \textbf{86.51 $\pm$ 3.49} \\
\bottomrule
\end{tabular}
\end{center}
\label{tab:multi-frequency}
\vspace{-6mm}
\end{table}

\begin{table}[h]
\caption{Emergent specialization in policy levels}
\vspace{-4mm}
\begin{center}
\begin{tabular}{l|cc}
\toprule
Level & \texttt{move-without-blowing} & \texttt{turn-while-blowing} \\
\midrule
High & 89.2\% & 10.8\% \\
Low & 17.7\% & 82.3\% \\
\bottomrule
\end{tabular}
\end{center}
\label{tab:emergent-specialization}
\vspace{-4mm}
\end{table}

\mysubsection{Three-level vs.\ two-level.} Our third experiment investigates whether multi-frequency policies with three levels are better than ones with two.
The results are shown for policies with $k=4$ at all levels in Tab.~\ref{tab:three-level}.
We find that the three-level policies generally perform slightly better, but require longer training times.

\begin{table}[h]
\vspace{-1mm}
\caption{Three-level vs.\ two-level}
\vspace{-4mm}
\begin{center}
\begin{tabular}{l|cc}
\toprule
Environment & 2-level & 3-level \\
\midrule
SmallEmpty & 48.52 $\pm$ 0.37 & \textbf{48.93 $\pm$ 0.33} \\
LargeEmpty & 93.09 $\pm$ 2.29 & \textbf{96.14 $\pm$ 2.03} \\
LargeColumns & 91.01 $\pm$ 3.58 & \textbf{95.06 $\pm$ 1.65} \\
LargeDoor & 89.58 $\pm$ 3.75 & \textbf{91.78 $\pm$ 2.63} \\
LargeCenter & \textbf{86.51 $\pm$ 3.49} & 84.15 $\pm$ 4.74 \\
\bottomrule
\end{tabular}
\end{center}
\label{tab:three-level}
\vspace{-4mm}
\end{table}

\begin{figure}
\begin{center}
\setlength\tabcolsep{1pt}
\begin{tabular}{cc}
\includegraphics[width=0.49\columnwidth]{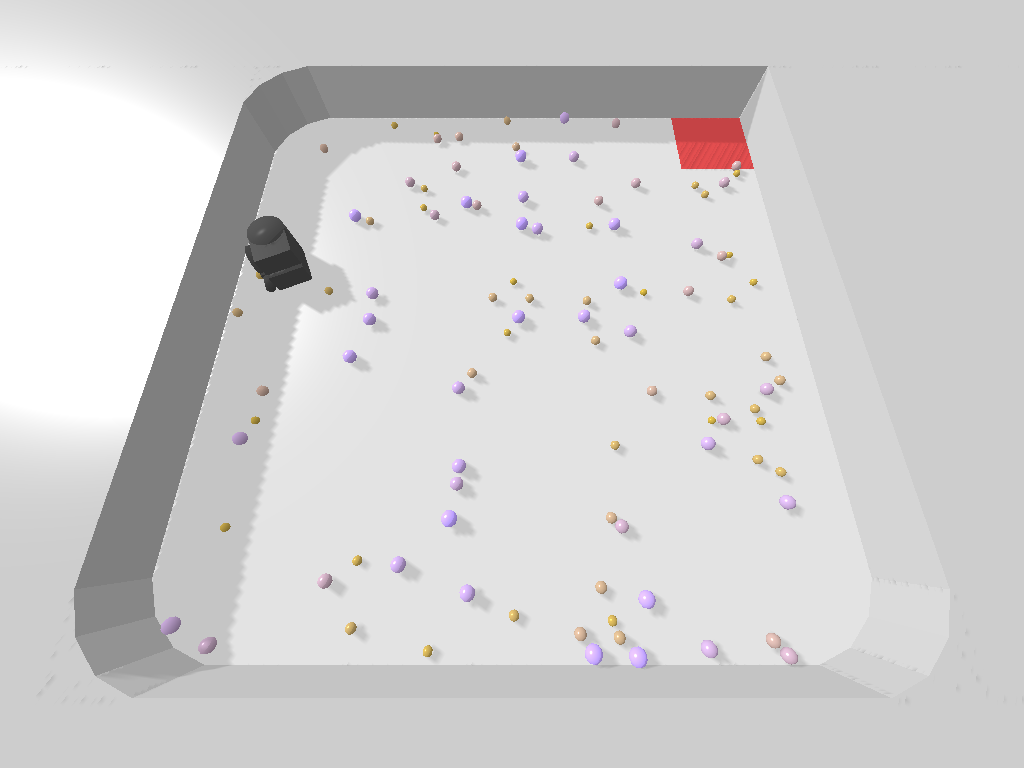} &
\includegraphics[width=0.49\columnwidth]{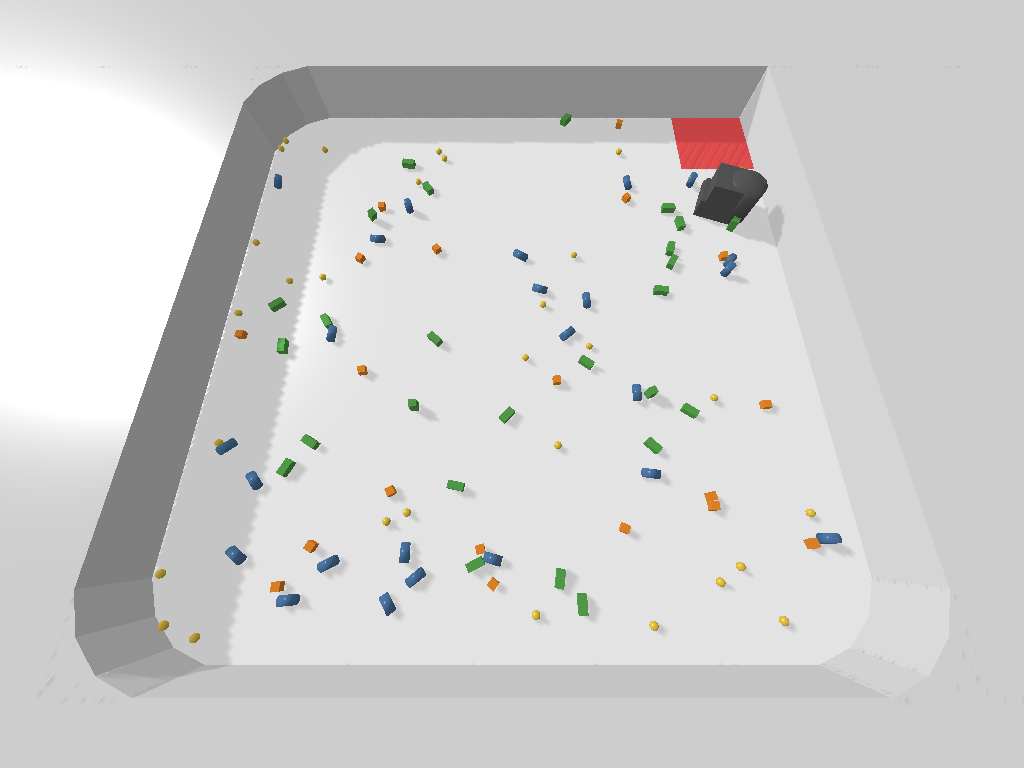} \\
\small{Objects of mixed sizes} & \small{Objects of mixed shapes} \\
\end{tabular}
\end{center}
\vspace{-2mm}
\caption{\textbf{Generalization to heterogeneous objects.} We study how well our policies generalize to two types of heterogeneous objects not seen during training: objects of mixed sizes (left) and objects of mixed shapes (right).}
\label{fig:heterogeneous-objects}
\vspace{-3mm}
\end{figure}

\mysubsection{Generalization to heterogeneous objects.} Our fourth experiment studies how well our policies can generalize to environments with heterogeneous objects. For each of our five environments, we create two heterogeneous variants as shown in Fig.~\ref{fig:heterogeneous-objects}.
The first uses spherical objects of various diameters with proportionally scaled masses.
The second uses a mixture of four different shapes: spheres, cubes, rectangular cuboids, and cylinders.
We directly test our policies on these unseen sets of objects with no fine-tuning.
Results are shown in Tab.~\ref{tab:generalization-heterogeneous}.
We find that our policies generalize well to heterogeneous objects of mixed sizes and shapes even though they were only trained with homogeneous spherical objects.

\begin{table}[h]
\caption{Generalization to heterogeneous objects}
\vspace{-4mm}
\begin{center}
\begin{tabular}{l|ccc}
\toprule
Environment & Reference & Mixed sizes & Mixed shapes \\
\midrule
SmallEmpty & 48.52 $\pm$ 0.37 & 49.48 $\pm$ 0.35 & 49.12 $\pm$ 0.28 \\
LargeEmpty & 93.09 $\pm$ 2.29 & 96.81 $\pm$ 0.92 & 95.12 $\pm$ 1.23 \\
LargeColumns & 91.01 $\pm$ 3.58 & 95.18 $\pm$ 2.43 & 93.98 $\pm$ 2.51 \\
LargeDoor & 89.58 $\pm$ 3.75 & 93.95 $\pm$ 2.06 & 87.32 $\pm$ 2.23 \\
LargeCenter & 86.51 $\pm$ 3.49 & 91.46 $\pm$ 1.91 & 91.57 $\pm$ 2.05 \\
\bottomrule
\end{tabular}
\end{center}
\label{tab:generalization-heterogeneous}
\vspace{-4mm}
\end{table}

\begin{figure}[t]
\begin{center}
\setlength\tabcolsep{1pt}
\begin{tabular}{cc}
\includegraphics[width=0.49\columnwidth]{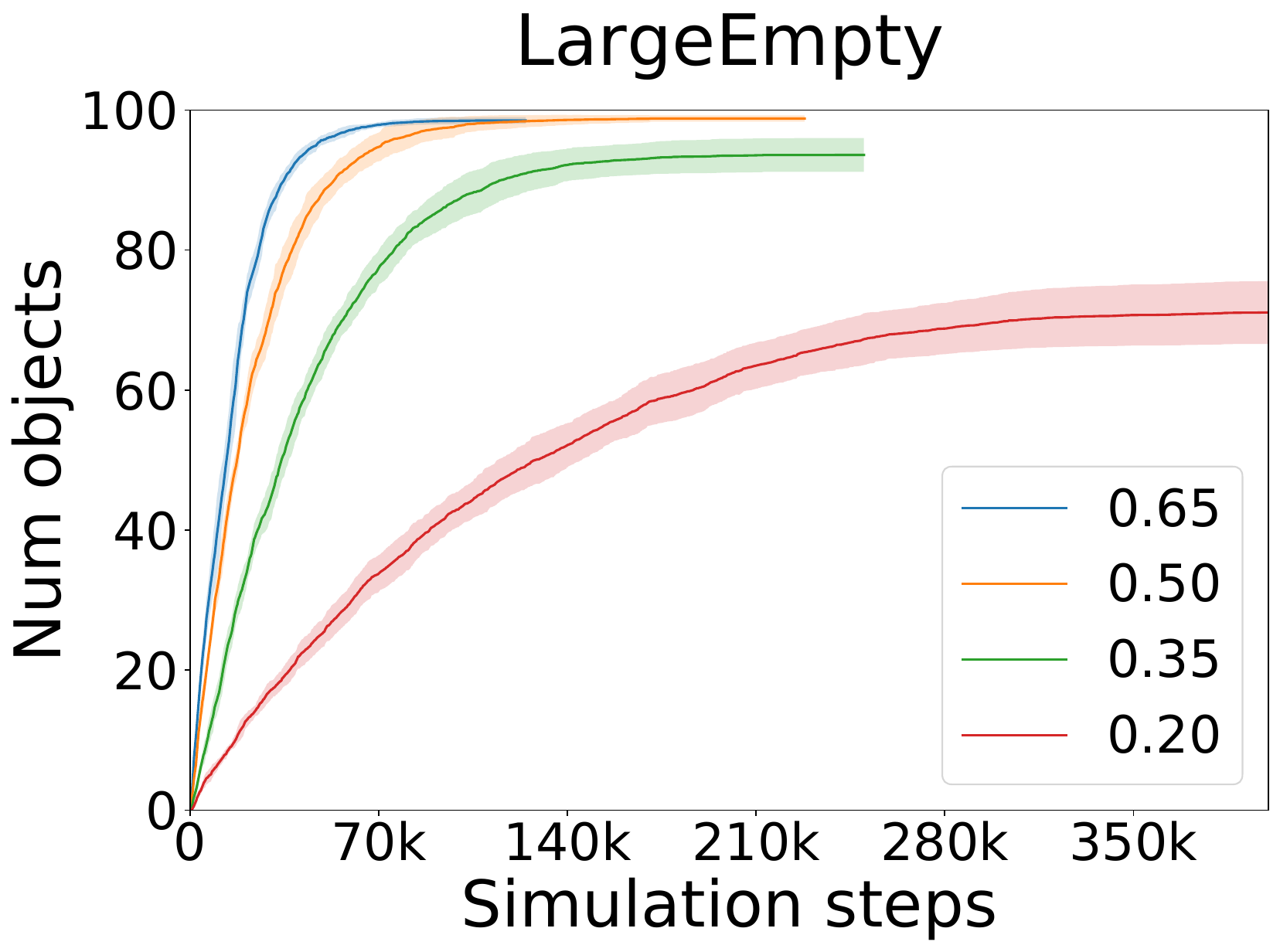} &
\includegraphics[width=0.49\columnwidth]{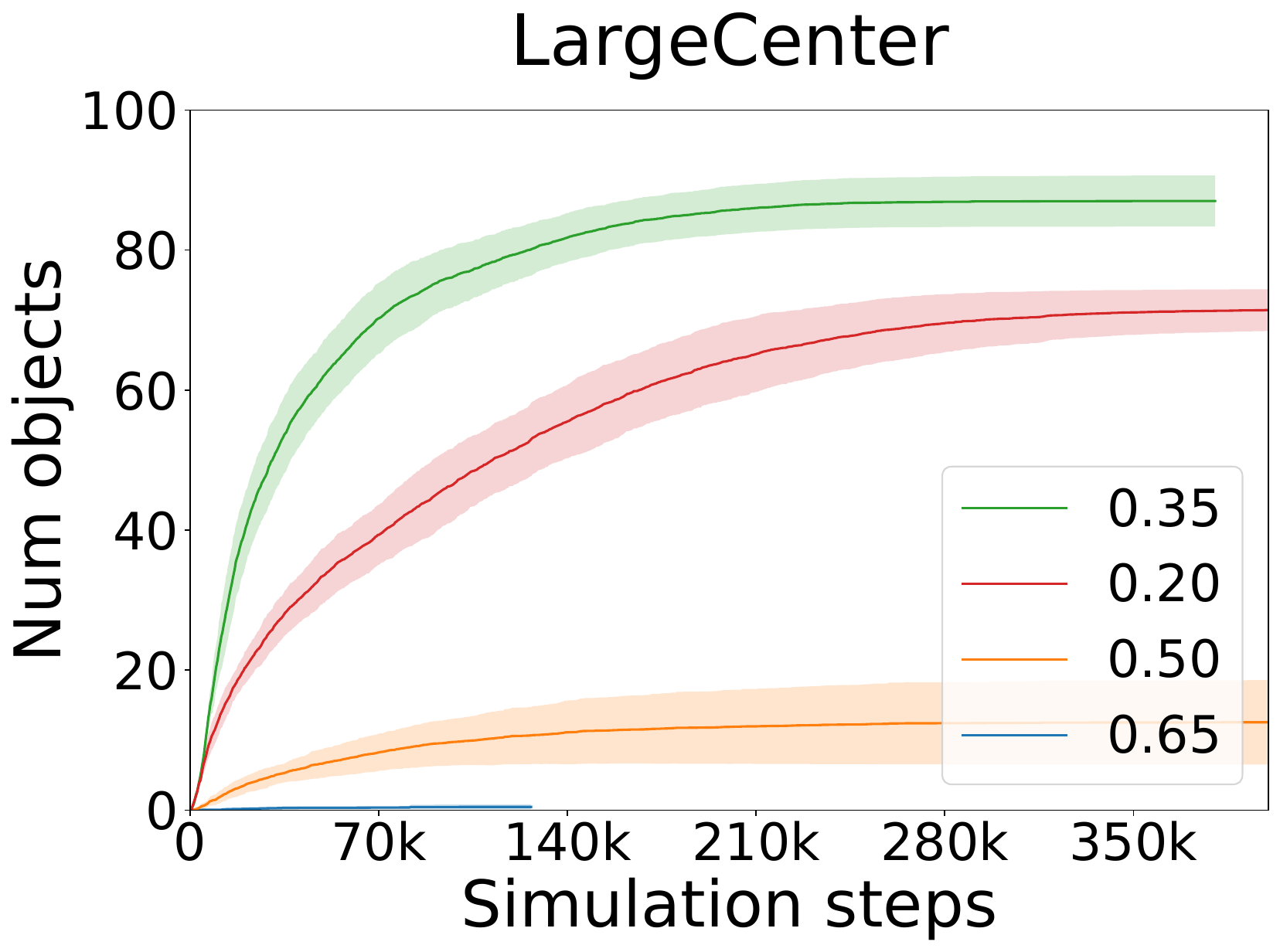} \\
\end{tabular}
\end{center}
\vspace{-3mm}
\caption{\textbf{Blowing force.}
We run an experiment where we vary the strength of the blower. We find that a stronger blower helps if the receptacle is in the corner, but hurts if the receptacle is in the center.}
\label{fig:blowing-force-curves}
\vspace{-3mm}
\end{figure}

\mysubsection{Blowing force.}
Our fifth experiment investigates the effect of blowing force magnitude on performance. By adjusting the force with which spherical ``wind'' particles are ejected from the blower, we can vary the strength of the blower. We train policies with different blowing forces ranging from 0.2 to 0.65 (where 0.35 is used in other experiments).
The results are shown in Fig.~\ref{fig:blowing-force-curves}.
We find that when the receptacle is in the corner (LargeEmpty), a stronger blower is able to complete the task more quickly since it can move the objects a longer distance with each action.
However, when the receptacle is in the center (LargeCenter), a blower that is too strong is unable to control the motion of objects well enough to get them into the receptacle.
A moderate strength (0.35) provides a nice balance, as shown in other experiments.
However, to get the best of both worlds, perhaps the policy could learn to vary the blowing strength in future work.

\mysubsection{Other blowing setups.}
Our sixth experiment looks at alternative blowing setups to study in what scenarios the multi-frequency approach helps.
We consider two variants: (i) a ``side blower'' robot, where the only difference from the base setup is that the blower points to the right side rather than forward, and (ii) a ``moving blower'' robot that is able to move with its blower on (\ie it uses the \texttt{move-while-blowing} action channel instead of \texttt{turn-while-blowing}).
The results are shown in Tab.~\ref{tab:other-blowing-setups}.
We find that in these alternative setups, the robot can still learn to do the task.
However, qualitatively, we observe that (i) the side blower struggles to move objects towards the center, and (ii) the moving blower exhibits poor distance efficiency.
Quantitatively, we find that multi-frequency consistently helps in more complex scenarios where there is a mismatch in execution frequencies, such as the environments in which the robot must strategically place itself before blowing (LargeColumns, LargeDoor, and LargeCenter).
We observe that multi-frequency does not help as much in simpler setups where strategic placement is not as important, such as the moving blower in SmallEmpty.

\begin{table}[h]
\caption{Other blowing setups}
\vspace{-4mm}
\begin{center}
\setlength\tabcolsep{0.34em}
\begin{tabular}{l|cc|cc}
\toprule
\multirow{2}{*}{Environment} & \multicolumn{2}{c|}{Side blower} & \multicolumn{2}{c}{Moving blower} \\
& Single-freq. & Multi-freq. & Single-freq. & Multi-freq. \\
\midrule
SmallEmpty & 43.03 $\!\pm\!$ \x2.41 & \textbf{46.62 $\!\pm\!$ 0.32} & \textbf{49.45 $\!\pm\!$ 0.27} & 49.05 $\!\pm\!$ 0.50 \\
LargeEmpty & 78.30 $\!\pm\!$ 10.25 & \textbf{89.40 $\!\pm\!$ 4.86} & 98.21 $\!\pm\!$ 1.19 & \textbf{98.42 $\!\pm\!$ 1.17} \\
LargeColumns & 81.35 $\!\pm\!$ \x2.51 & \textbf{91.53 $\!\pm\!$ 1.31} & 92.90 $\!\pm\!$ 2.68 & \textbf{97.03 $\!\pm\!$ 1.81} \\
LargeDoor & 60.73 $\!\pm\!$ 12.43 & \textbf{74.99 $\!\pm\!$ 6.64} & 88.55 $\!\pm\!$ 5.99 & \textbf{91.77 $\!\pm\!$ 1.86} \\
LargeCenter & 52.73 $\!\pm\!$ \x3.45 & \textbf{63.46 $\!\pm\!$ 1.53} & 84.81 $\!\pm\!$ 8.05 & \textbf{88.83 $\!\pm\!$ 2.82} \\
\bottomrule
\end{tabular}
\end{center}
\label{tab:other-blowing-setups}
\vspace{-4mm}
\end{table}

\subsection{Real-World Experiments}

Compared to the simple blowing model we use in simulation (spherical ``wind'' particles), the aerodynamics of real-world blowing are much more complex. To test how well our policies can generalize to real-world dynamics, we run our best policies from simulation on a real robot.
We use an Anki Vector robot equipped with a simple forward-facing battery-powered air blower attached to its articulated arm (Fig. \ref{fig:teaser}). We place the robot into an environment with orange colored spheres and a square region serving as the receptacle.

To handle the sim-to-real transfer, we mimic the real-world setup in simulation (for the SmallEmpty environment), with the simulated robot and blower calibrated against the real ones.
We use fiducial markers to track the robot's pose, which we mirror inside the simulation.
For the scattered objects, we use an overhead camera along with simple thresholding in HSV color space to generate a segmentation mask of the objects.
This mask is used to generate an appropriate observation inside the simulation for the agent's simulated online mapping.
Note that no pose estimation for the objects is necessary, as the policy is trained on an image-based state representation.
We execute the simulation-trained policies directly on the real robot with no fine-tuning.

\begin{figure}
\begin{center}
\includegraphics[width=0.325\columnwidth]{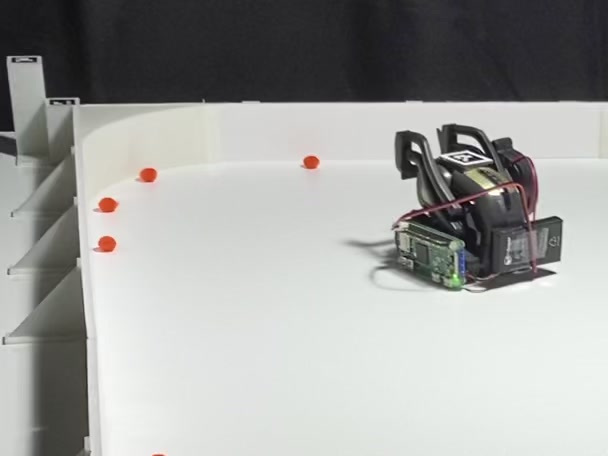}
\includegraphics[width=0.325\columnwidth]{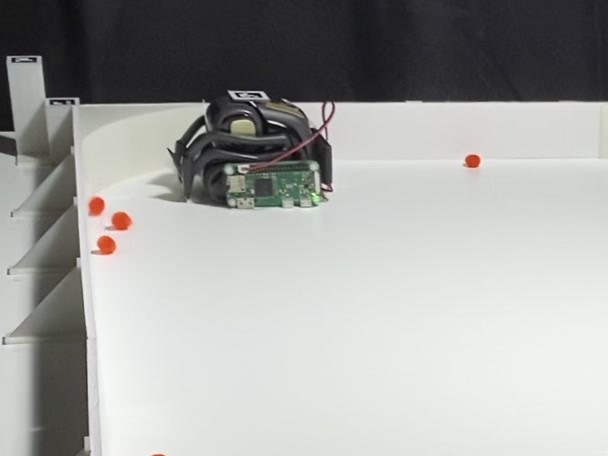}
\includegraphics[width=0.325\columnwidth]{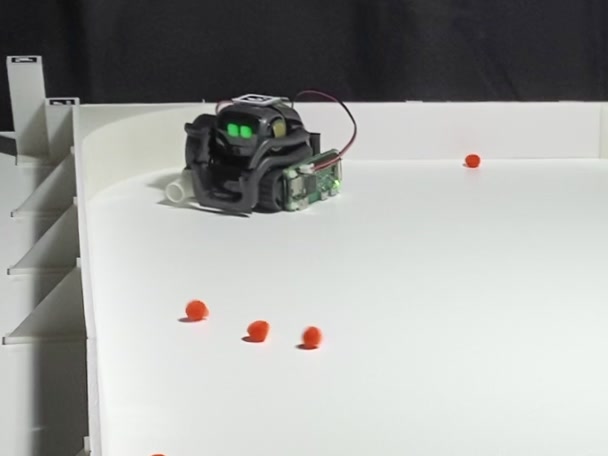}

\vspace{-1mm}
\caption{\textbf{Emergent behavior.} We observe a strategy in which the robot blows into a corner and uses the reflected air flow to move objects out. This behavior was learned in simulation, and transfers successfully to the real robot without fine-tuning.}
\label{fig:emergent-behavior}
\vspace{-3mm}
\end{center}
\end{figure}

\mysubsection{Qualitative results.}
Fig.~\ref{fig:emergent-behavior} shows an example result of running the real robot in a real-world replica of the SmallEmpty environment.  The figure illustrates how one of the interesting emergent behaviors learned in simulation can transfer to the real world: by blowing against a wall while turning, the robot is able to dislodge objects from corners.
For qualitative video results and code, please see our supplementary material at \url{https://learning-dynamic-manipulation.cs.princeton.edu}.

\begin{figure}
\begin{center}
\setlength\tabcolsep{1pt}
\begin{tabular}{cc}
\includegraphics[width=0.49\columnwidth]{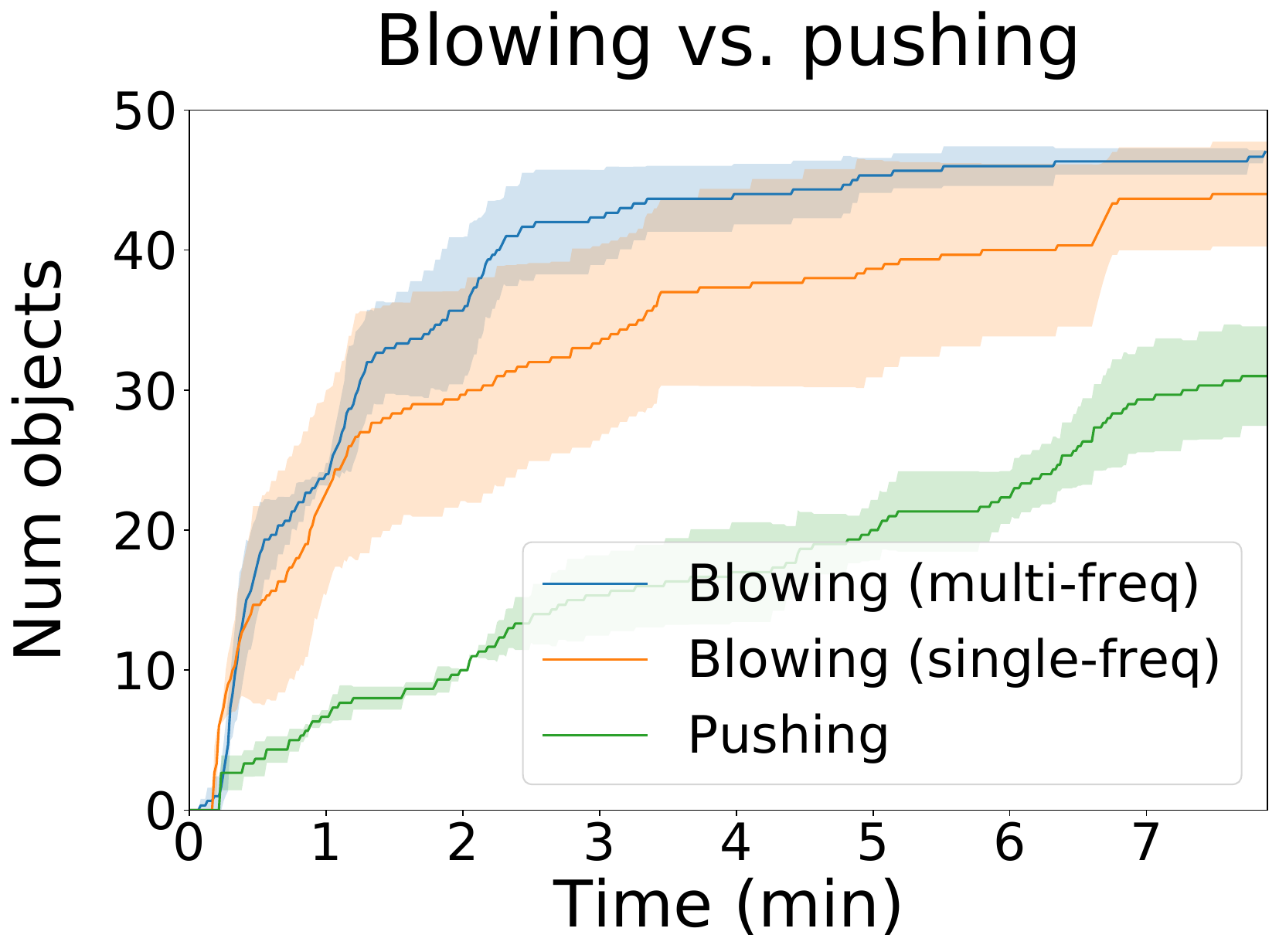} &
\includegraphics[width=0.49\columnwidth]{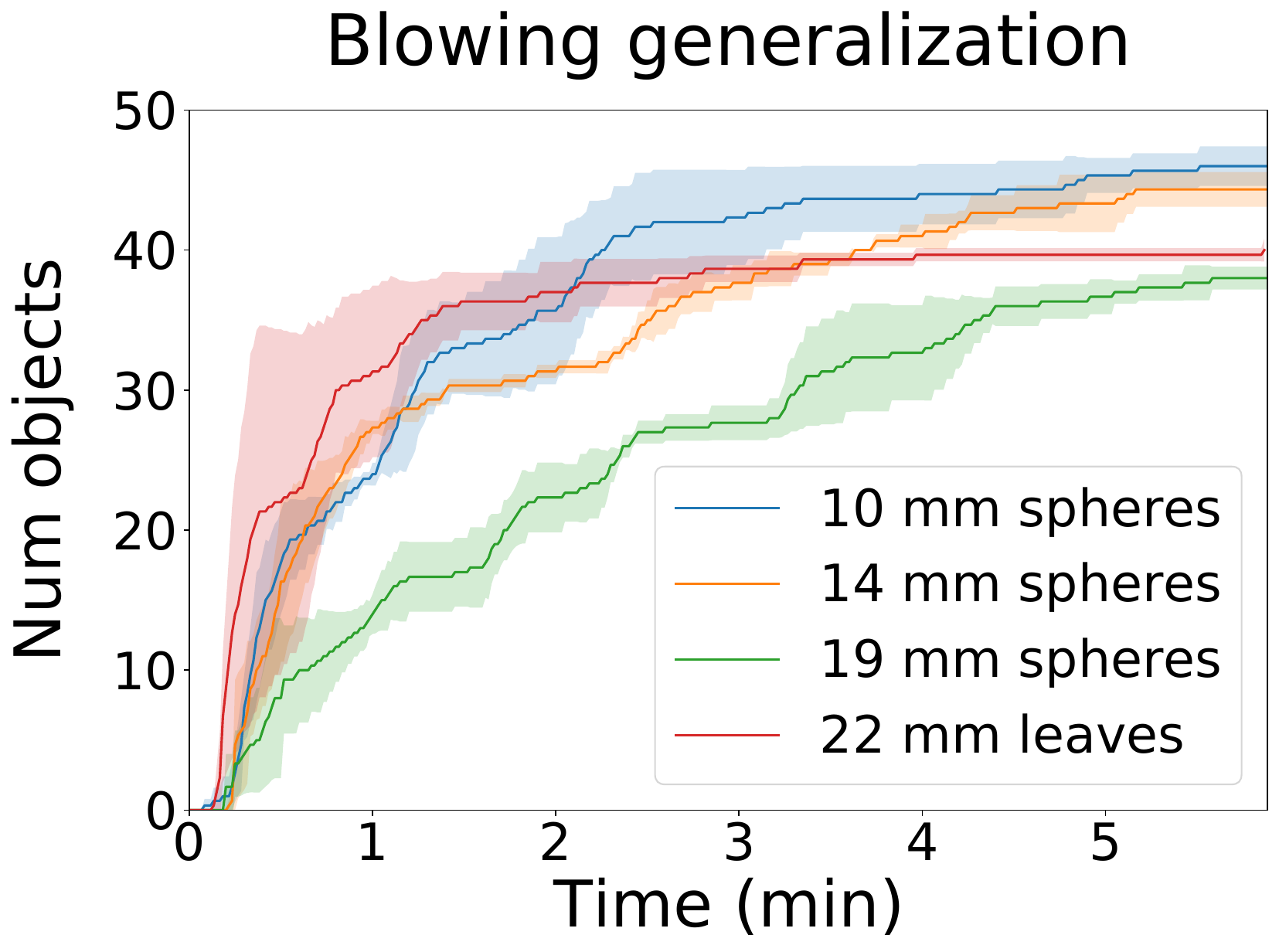} \\
\end{tabular}
\end{center}
\vspace{-3mm}
\caption{\textbf{Real-world experiments.} Here we show evaluation curves for our real-world experiments. On the left, we compare blowing and pushing. On the right, we look at how well our policy generalizes to novel objects.}
\label{fig:real-robot-curves}
\vspace{-3mm}
\end{figure}

\mysubsection{Blowing vs.\ pushing.} We ran an experiment to perform a quantitative comparison between blowing and pushing in the real world.   Specifically, we ran the blowing and pushing policies in the real-world replica of the SmallEmpty environment and recorded the number of objects moved into the receptacle as a function of elapsed time.
We ran 3 evaluation episodes for each policy and show the mean and standard deviation across the episodes in the left plot of Fig.~\ref{fig:real-robot-curves}.
The results are comparable to those for SmallEmpty obtained in simulation in Fig.~\ref{fig:blowing-pushing-curves}.
The multi-frequency blowing policy is able to get over 40 objects by the 3 minute mark, whereas the pushing policy has gotten less than 20 by that point.

\begin{figure}
\begin{center}
\setlength\tabcolsep{1pt}
\begin{tabular}{cccc}
\includegraphics[width=0.24\columnwidth]{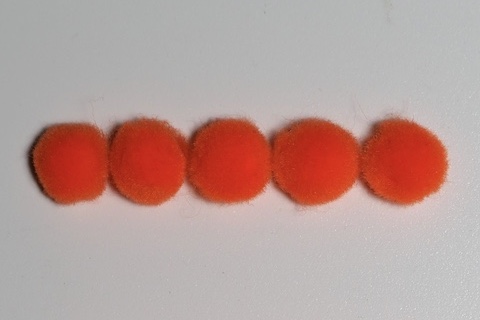} &
\includegraphics[width=0.24\columnwidth]{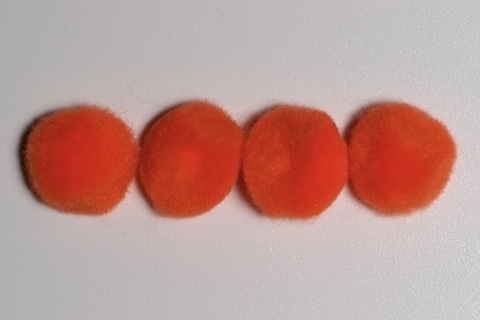} &
\includegraphics[width=0.24\columnwidth]{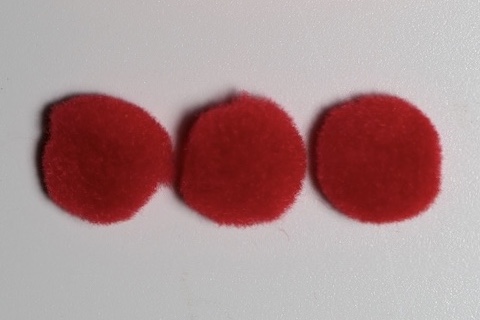} &
\includegraphics[width=0.24\columnwidth]{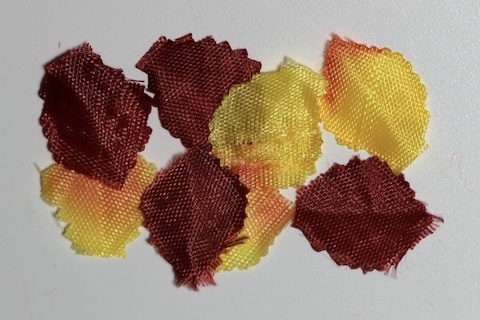} \\
\small{10\,mm spheres} & \small{14\,mm spheres} & \small{19\,mm spheres} & \small{22\,mm leaves} \\
\end{tabular}
\vspace{-2mm}
\caption{\textbf{Real-world objects.}
We test how well our trained policy generalizes to novel objects with different sizes and shapes. Our policies were trained in simulation with 10\,mm spherical objects (left), and tested in the real environment with the other objects shown.}
\label{fig:generalization-objects}
\vspace{-3mm}
\end{center}
\end{figure}

\mysubsection{Generalization to novel objects.} We also investigate whether policies trained in simulation on one type of object (10\,mm spheres) can generalize (with no fine-tuning) to real-world objects of different sizes and shapes.
Specifically, we test our policy on spherical objects of different sizes and leaves with irregular shapes (see Fig.~\ref{fig:generalization-objects}).
For each novel object, the only change we make to the system is adjustment of HSV thresholding values to accommodate the object's color.
Since our policy takes in an image-based state representation, policies trained on one object can be directly used for other objects without retraining.

The results of the experiment are in the right plot of Fig.~\ref{fig:real-robot-curves}.
When testing with larger objects, we find that the task takes longer to complete.
Since the larger objects are heavier, they do not travel as far when blown, so they require more actions to be moved the same distance.
However, the system is still able to complete the task for all of the novel objects.
The system is even able to generalize well to loose maple leaves.
Due to their light weight, we find that the performance is very good at the beginning of each episode since the leaves move a very large distance when blown.
However, some leaves are blown out of the environment, which is why the curve flattens out.

\section{Conclusion}

In this work, we present a mobile robot system that uses pneumatic blowing, a form of dynamic non-prehensile manipulation, to efficiently move scattered objects into a target receptacle.
Pneumatic mobile manipulation presents new challenges in dealing with the chaotic nature of aerodynamic forces: from requiring high-level spatial reasoning, to maintaining low-level fine-grained control.
We find that our proposed multi-frequency strategy not only outperforms the single-frequency baseline, but also results in naturally emerging specialization and collaboration between different levels of subpolicies.
Additionally, we find that our simulation-trained policies generalize well to the complex dynamics of blowing in the real world, and even to novel objects of different sizes and shapes.

\section*{Acknowledgments}
We thank Naomi Leonard, Anirudha Majumdar, Naveen Verma, Yen-Chen Lin, Kevin Zakka, and Rohan Agrawal for fruitful technical discussions. This work was supported by Princeton School of Engineering, as well as NSF under IIS-1815070 and DGE-1656466.

\bibliographystyle{IEEEtran}
\bibliography{IEEEabrv,references}

\begin{thebibliography}{10}
\providecommand{\url}[1]{#1}
\csname url@rmstyle\endcsname
\providecommand{\newblock}{\relax}
\providecommand{\bibinfo}[2]{#2}
\providecommand\BIBentrySTDinterwordspacing{\spaceskip=0pt\relax}
\providecommand\BIBentryALTinterwordstretchfactor{4}
\providecommand\BIBentryALTinterwordspacing{\spaceskip=\fontdimen2\font plus
\BIBentryALTinterwordstretchfactor\fontdimen3\font minus
  \fontdimen4\font\relax}
\providecommand\BIBforeignlanguage[2]{{%
\expandafter\ifx\csname l@#1\endcsname\relax
\typeout{** WARNING: IEEEtran.bst: No hyphenation pattern has been}%
\typeout{** loaded for the language `#1'. Using the pattern for}%
\typeout{** the default language instead.}%
\else
\language=\csname l@#1\endcsname
\fi
#2}}

\bibitem{laurent2015survey}
G.~J. Laurent and H.~Moon, ``A survey of non-prehensile pneumatic manipulation
  surfaces: principles, models and control,'' \emph{Intelligent Service
  Robotics}, 2015.

\bibitem{becker2009automated}
A.~Becker, R.~Sandheinrich, and T.~Bretl, ``Automated manipulation of spherical
  objects in three dimensions using a gimbaled air jet,'' in \emph{2009
  IEEE/RSJ International Conference on Intelligent Robots and Systems}, 2009.

\bibitem{davis2008end}
S.~Davis, J.~Gray, and D.~G. Caldwell, ``An end effector based on the bernoulli
  principle for handling sliced fruit and vegetables,'' \emph{Robotics and
  Computer-Integrated Manufacturing}, 2008.

\bibitem{erzincanli1998design}
F.~Erzincanli, J.~Sharp, and S.~Erhal, ``Design and operational considerations
  of a non-contact robotic handling system for non-rigid materials,''
  \emph{International Journal of Machine Tools and Manufacture}, 1998.

\bibitem{ozcelik2002non}
B.~Ozcelik and F.~Erzincanli, ``A non-contact end-effector for the handling of
  garments,'' \emph{Robotica}, 2002.

\bibitem{wu2020spatial}
J.~Wu, X.~Sun, A.~Zeng, S.~Song, J.~Lee, S.~Rusinkiewicz, and T.~Funkhouser,
  ``Spatial action maps for mobile manipulation,'' in \emph{Proceedings of
  Robotics: Science and Systems (RSS)}, 2020.

\bibitem{lynch2003control}
K.~M. Lynch and T.~D. Murphey, ``Control of nonprehensile manipulation,'' in
  \emph{Control problems in robotics}, 2003.

\bibitem{mason1999progress}
M.~T. Mason, ``Progress in nonprehensile manipulation,'' \emph{The
  International Journal of Robotics Research}, 1999.

\bibitem{mason1993dynamic}
M.~T. Mason and K.~M. Lynch, ``Dynamic manipulation,'' in \emph{Proceedings of
  1993 IEEE/RSJ International Conference on Intelligent Robots and Systems
  (IROS'93)}, 1993.

\bibitem{wang2020swingbot}
C.~Wang, S.~Wang, B.~Romero, F.~Veiga, and E.~Adelson, ``Swingbot: Learning
  physical features from in-hand tactile exploration for dynamic swing-up
  manipulation,'' in \emph{2020 IEEE/RSJ International Conference on
  Intelligent Robots and Systems (IROS)}, 2020.

\bibitem{zeng2020tossingbot}
A.~Zeng, S.~Song, J.~Lee, A.~Rodriguez, and T.~Funkhouser, ``Tossingbot:
  Learning to throw arbitrary objects with residual physics,'' \emph{IEEE
  Transactions on Robotics}, 2020.

\bibitem{ha2021flingbot}
H.~Ha and S.~Song, ``Flingbot: The unreasonable effectiveness of dynamic
  manipulation for cloth unfolding,'' \emph{arXiv preprint arXiv:2105.03655},
  2021.

\bibitem{xu2022dextairity}
Z.~Xu, C.~Chi, B.~Burchfiel, E.~Cousineau, S.~Feng, and S.~Song, ``Dextairity:
  Deformable manipulation can be a breeze,'' \emph{arXiv preprint
  arXiv:2203.01197}, 2022.

\bibitem{konishi1999development}
S.~Konishi, Y.~Mizuguchi, and K.~Ohno, ``Development of a non-contact
  conveyance system composed of distributed nozzle units,'' in \emph{1999 7th
  IEEE International Conference on Emerging Technologies and Factory
  Automation}, 1999.

\bibitem{reed2004high}
J.~Reed and S.~Miles, ``High-speed conveyor junction based on an air-jet
  floatation technique,'' \emph{Mechatronics}, 2004.

\bibitem{luntz2001distributed}
J.~Luntz and H.~Moon, ``Distributed manipulation with passive air flow,'' in
  \emph{Proceedings 2001 IEEE/RSJ International Conference on Intelligent
  Robots and Systems. Expanding the Societal Role of Robotics in the the Next
  Millennium (Cat. No. 01CH37180)}, 2001.

\bibitem{escano2005position}
J.~M. Escano, M.~G. Ortega, and F.~R. Rubio, ``Position control of a pneumatic
  levitation system,'' in \emph{2005 IEEE Conference on Emerging Technologies
  and Factory Automation}, 2005.

\bibitem{nordine1982aerodynamic}
P.~C. Nordine and R.~M. Atkins, ``Aerodynamic levitation of laser-heated solids
  in gas jets,'' \emph{Review of Scientific Instruments}, 1982.

\bibitem{ozcelik2005examination}
B.~Ozcelik and F.~Erzincanli, ``Examination of the movement of a woven fabric
  in the horizontal direction using a non-contact end-effector,'' \emph{The
  International Journal of Advanced Manufacturing Technology}, 2005.

\bibitem{biegelsen2000airjet}
D.~Biegelsen, A.~Berlin, P.~Cheung, M.~Fromherz, D.~Goldberg, W.~Jackson,
  B.~Preas, J.~Reich, and L.~Swartz, ``Airjet paper mover,'' in \emph{Presented
  at SPIE Int. Symposium on Micromachining and Microfabrication}, 2000.

\bibitem{ijspeert2013dynamical}
A.~J. Ijspeert, J.~Nakanishi, H.~Hoffmann, P.~Pastor, and S.~Schaal,
  ``Dynamical movement primitives: learning attractor models for motor
  behaviors,'' \emph{Neural computation}, 2013.

\bibitem{mulling2013learning}
K.~M{\"u}lling, J.~Kober, O.~Kroemer, and J.~Peters, ``Learning to select and
  generalize striking movements in robot table tennis,'' \emph{The
  International Journal of Robotics Research}, 2013.

\bibitem{saveriano2021dynamic}
M.~Saveriano, F.~J. Abu-Dakka, A.~Kramberger, and L.~Peternel, ``Dynamic
  movement primitives in robotics: A tutorial survey,'' \emph{arXiv preprint
  arXiv:2102.03861}, 2021.

\bibitem{daniel2012hierarchical}
C.~Daniel, G.~Neumann, and J.~Peters, ``Hierarchical relative entropy policy
  search,'' in \emph{Artificial Intelligence and Statistics}, 2012.

\bibitem{stulp2012reinforcement}
F.~Stulp, E.~A. Theodorou, and S.~Schaal, ``Reinforcement learning with
  sequences of motion primitives for robust manipulation,'' \emph{IEEE
  Transactions on robotics}, 2012.

\bibitem{sutton1999between}
R.~S. Sutton, D.~Precup, and S.~Singh, ``Between mdps and semi-mdps: A
  framework for temporal abstraction in reinforcement learning,''
  \emph{Artificial intelligence}, 1999.

\bibitem{tessler2017deep}
C.~Tessler, S.~Givony, T.~Zahavy, D.~Mankowitz, and S.~Mannor, ``A deep
  hierarchical approach to lifelong learning in minecraft,'' in
  \emph{Proceedings of the AAAI Conference on Artificial Intelligence}, 2017.

\bibitem{heess2016learning}
N.~Heess, G.~Wayne, Y.~Tassa, T.~Lillicrap, M.~Riedmiller, and D.~Silver,
  ``Learning and transfer of modulated locomotor controllers,'' \emph{arXiv
  preprint arXiv:1610.05182}, 2016.

\bibitem{haarnoja2018latent}
T.~Haarnoja, K.~Hartikainen, P.~Abbeel, and S.~Levine, ``Latent space policies
  for hierarchical reinforcement learning,'' in \emph{International Conference
  on Machine Learning}, 2018.

\bibitem{wang2021hierarchical}
L.~Wang, Y.~Xiang, and D.~Fox, ``Hierarchical policies for cluttered-scene
  grasping with latent plans,'' \emph{arXiv preprint arXiv:2107.01518}, 2021.

\bibitem{schaal2006dynamic}
S.~Schaal, ``Dynamic movement primitives-a framework for motor control in
  humans and humanoid robotics,'' in \emph{Adaptive motion of animals and
  machines}, 2006.

\bibitem{stulp2011hierarchical}
F.~Stulp and S.~Schaal, ``Hierarchical reinforcement learning with movement
  primitives,'' in \emph{2011 11th IEEE-RAS International Conference on
  Humanoid Robots}, 2011.

\bibitem{bahl2020neural}
S.~Bahl, M.~Mukadam, A.~Gupta, and D.~Pathak, ``Neural dynamic policies for
  end-to-end sensorimotor learning,'' in \emph{Advances in Neural Information
  Processing Systems}, 2020.

\bibitem{bahl2021hierarchical}
S.~Bahl, A.~Gupta, and D.~Pathak, ``Hierarchical neural dynamic policies,'' in
  \emph{Proceedings of Robotics: Science and Systems (RSS)}, 2021.

\bibitem{kulkarni2016hierarchical}
T.~D. Kulkarni, K.~Narasimhan, A.~Saeedi, and J.~Tenenbaum, ``Hierarchical deep
  reinforcement learning: Integrating temporal abstraction and intrinsic
  motivation,'' \emph{Advances in neural information processing systems}, 2016.

\bibitem{peng2017deeploco}
X.~B. Peng, G.~Berseth, K.~Yin, and M.~Van De~Panne, ``Deeploco: Dynamic
  locomotion skills using hierarchical deep reinforcement learning,'' \emph{ACM
  Transactions on Graphics (TOG)}, 2017.

\bibitem{faust2018prm}
A.~Faust, K.~Oslund, O.~Ramirez, A.~Francis, L.~Tapia, M.~Fiser, and
  J.~Davidson, ``Prm-rl: Long-range robotic navigation tasks by combining
  reinforcement learning and sampling-based planning,'' in \emph{2018 IEEE
  International Conference on Robotics and Automation (ICRA)}, 2018.

\bibitem{wahid2019long}
A.~Wahid, A.~Toshev, M.~Fiser, and T.-W.~E. Lee, ``Long range neural navigation
  policies for the real world,'' in \emph{2019 IEEE/RSJ International
  Conference on Intelligent Robots and Systems (IROS)}, 2019.

\bibitem{wang2020model}
R.~E. Wang, J.~C. Kew, D.~Lee, T.-W.~E. Lee, T.~Zhang, B.~Ichter, J.~Tan, and
  A.~Faust, ``Model-based reinforcement learning for decentralized multiagent
  rendezvous,'' in \emph{Proceedings of the Conference on Robot Learning},
  2020.

\bibitem{nachum2020multi}
O.~Nachum, M.~Ahn, H.~Ponte, S.~S. Gu, and V.~Kumar, ``Multi-agent manipulation
  via locomotion using hierarchical sim2real,'' in \emph{Proceedings of the
  Conference on Robot Learning}, 2020.

\bibitem{li2020hrl4in}
C.~Li, F.~Xia, R.~Martin-Martin, and S.~Savarese, ``Hrl4in: Hierarchical
  reinforcement learning for interactive navigation with mobile manipulators,''
  in \emph{Conference on Robot Learning}, 2020.

\bibitem{xiali2020relmogen}
F.~Xia, C.~Li, R.~Mart{\'\i}n-Mart{\'\i}n, O.~Litany, A.~Toshev, and
  S.~Savarese, ``Relmogen: Leveraging motion generation in reinforcement
  learning for mobile manipulation,'' in \emph{Conference on Robot Learning},
  2021.

\bibitem{stolle2002learning}
M.~Stolle and D.~Precup, ``Learning options in reinforcement learning,'' in
  \emph{International Symposium on abstraction, reformulation, and
  approximation}, 2002.

\bibitem{levy2017learning}
A.~Levy, G.~Konidaris, R.~Platt, and K.~Saenko, ``Learning multi-level
  hierarchies with hindsight,'' in \emph{International Conference on Learning
  Representations}, 2019.

\bibitem{nachum2018data}
O.~Nachum, S.~S. Gu, H.~Lee, and S.~Levine, ``Data-efficient hierarchical
  reinforcement learning,'' in \emph{Advances in Neural Information Processing
  Systems}, 2018.

\bibitem{morimoto2001acquisition}
J.~Morimoto and K.~Doya, ``Acquisition of stand-up behavior by a real robot
  using hierarchical reinforcement learning,'' \emph{Robotics and Autonomous
  Systems}, 2001.

\bibitem{vezhnevets2017feudal}
A.~S. Vezhnevets, S.~Osindero, T.~Schaul, N.~Heess, M.~Jaderberg, D.~Silver,
  and K.~Kavukcuoglu, ``Feudal networks for hierarchical reinforcement
  learning,'' in \emph{International Conference on Machine Learning}, 2017.

\bibitem{florensa2017stochastic}
C.~Florensa, Y.~Duan, and P.~Abbeel, ``Stochastic neural networks for
  hierarchical reinforcement learning,'' in \emph{International Conference on
  Learning Representations}, 2017.

\bibitem{camacho2021reward}
A.~Camacho, J.~Varley, A.~Zeng, D.~Jain, A.~Iscen, and D.~Kalashnikov, ``Reward
  machines for vision-based robotic manipulation,'' in \emph{2021 IEEE
  International Conference on Robotics and Automation (ICRA)}, 2021.

\bibitem{frans2017meta}
K.~Frans, J.~Ho, X.~Chen, P.~Abbeel, and J.~Schulman, ``Meta learning shared
  hierarchies,'' in \emph{International Conference on Learning
  Representations}, 2018.

\bibitem{sun2022fully}
C.~Sun, J.~Orbik, C.~M. Devin, B.~H. Yang, A.~Gupta, G.~Berseth, and S.~Levine,
  ``Fully autonomous real-world reinforcement learning with applications to
  mobile manipulation,'' in \emph{Conference on Robot Learning}, 2022.

\bibitem{eysenbach2018diversity}
B.~Eysenbach, A.~Gupta, J.~Ibarz, and S.~Levine, ``Diversity is all you need:
  Learning skills without a reward function,'' in \emph{International
  Conference on Learning Representations}, 2019.

\bibitem{mnih2015human}
V.~Mnih, K.~Kavukcuoglu, D.~Silver, A.~A. Rusu, J.~Veness, M.~G. Bellemare,
  A.~Graves, M.~Riedmiller, A.~K. Fidjeland, G.~Ostrovski, \emph{et~al.},
  ``Human-level control through deep reinforcement learning,'' \emph{Nature},
  2015.

\bibitem{van2016deep}
H.~Van~Hasselt, A.~Guez, and D.~Silver, ``Deep reinforcement learning with
  double q-learning,'' in \emph{Thirtieth AAAI conference on artificial
  intelligence}, 2016.

\bibitem{coumans2021pybullet}
E.~Coumans and Y.~Bai, ``Pybullet, a python module for physics simulation for
  games, robotics and machine learning,'' \url{http://pybullet.org},
  2016--2021.

\bibitem{he2016deep}
K.~He, X.~Zhang, S.~Ren, and J.~Sun, ``Deep residual learning for image
  recognition,'' in \emph{IEEE Conference on Computer Vision and Pattern
  Recognition}, 2016.

\bibitem{long2015fully}
J.~Long, E.~Shelhamer, and T.~Darrell, ``Fully convolutional networks for
  semantic segmentation,'' in \emph{IEEE conference on computer vision and
  pattern recognition}, 2015.

\end{thebibliography}

\end{document}